\documentclass[letterpaper, 10 pt, journal, twoside]{IEEEtran}  %

\IEEEoverridecommandlockouts                              %
\usepackage[dvipsnames]{xcolor}
\usepackage[export]{adjustbox}
\usepackage{graphicx}
\usepackage{amsmath}
\usepackage{amssymb}
\usepackage{booktabs}
\usepackage{makecell}
\usepackage{gensymb}
\usepackage{siunitx}
\usepackage{csquotes}
\usepackage{microtype}

\usepackage{enumitem}
\usepackage[caption=false]{subfig}
\usepackage{pgfplots}
\usepackage{cite}
\usepackage{flushend}
\usepackage{ragged2e}
\usepackage[hang,flushmargin]{footmisc}
\pgfplotsset{compat=1.16}

\DeclareMathOperator{\avg}{avg}

\usepackage{pifont}%

\usepackage[breaklinks,colorlinks,bookmarks=false]{hyperref}
\usepackage[capitalize]{cleveref}
\crefname{section}{Sec.}{Secs.}
\Crefname{section}{Section}{Sections}
\Crefname{table}{Table}{Tables}
\crefname{table}{Tab.}{Tabs.}

\definecolor{dark-green}{RGB}{12,80,12}

\newcommand{\para}[1]{\parskip=5pt\noindent\textit{#1}}

\definecolor{Yellow}{RGB}{226,155,36}

\usepackage{threeparttable}
\usepackage{tabularx}
\newcolumntype{Y}{>{\centering\arraybackslash}X}
\newcolumntype{Z}{>{\raggedleft\arraybackslash}X}
\newcolumntype{R}{>{\raggedright\arraybackslash}X}

\usepackage[resetlabels]{multibib}
\newcites{S}{References}
\usepackage{cuted}  %

\usepackage{utfsym}
\newcommand{\yes}{\large \color{OliveGreen}\checkmark}
\newcommand{\no}{\color{BrickRed} \scalebox{1}{\usym{2613}}}
\usepackage{multirow}

\usepackage{varwidth}

\newcommand{\ourName}{\mbox{MoMa-Teleop}}

\newcommand{\website}{\url{https://moma-teleop.cs.uni-freiburg.de}}

\newcommand{\myworries}[1]{{\color{black}{#1}}}
\colorlet{myworriestablecolor}{black}
\newcommand{\myworriestwo}[1]{{\color{black}{#1}}}
\newcommand{\ntwo}{N$^2$M$^2$}

\newcommand{\ourtitle}{\myworriestwo{Whole-Body Teleoperation for Mobile Manipulation at Zero Added Cost}}

\title{
\ourtitle{}
}

\author{
Daniel Honerkamp$^{*}$, Harsh Mahesheka$^{*}$, Jan Ole von Hartz, Tim Welschehold and Abhinav Valada%
\thanks{$^*$Equal contribution. All authors are with the Department of Computer Science, University of Freiburg, Germany.}%
\thanks{This work was partially funded by the German Research Foundation (DFG): 417962828, an academic grant from NVIDIA, the BrainLinks-BrainTools center of the University of Freiburg, and supported with an HSR robot by Toyota Motor Europe.}%
\thanks{© 2025 IEEE.  Personal use of this material is permitted.  Permission from IEEE must be obtained for all other uses, in any current or future media, including reprinting/republishing this material for advertising or promotional purposes, creating new collective works, for resale or redistribution to servers or lists, or reuse of any copyrighted component of this work in other works.}%
}

\begin{document}

\maketitle
\begin{abstract}

Demonstration data plays a key role in learning complex behaviors and training robotic foundation models. While effective control interfaces exist for static manipulators, data collection remains cumbersome and time intensive for mobile manipulators due to their large number of degrees of freedom. While specialized hardware, avatars, or motion tracking can enable whole-body control, these approaches are either expensive, robot-specific, or suffer from the embodiment mismatch between robot and human demonstrator. In this work, we present \ourName{}, a novel teleoperation method that \myworries{infers end-effector motions from existing interfaces} and delegates the base motions to a \myworries{previously developed} reinforcement learning agent, leaving the operator to focus fully on the task-relevant end-effector motions. This enables whole-body teleoperation of mobile manipulators with \myworries{no} additional hardware or setup costs via standard interfaces such as joysticks or hand guidance. Moreover, the operator is not bound to a tracked workspace and can move freely with the robot over spatially extended tasks. We demonstrate that our approach results in a significant reduction in task completion time across a variety of robots and tasks. As the generated data covers diverse whole-body motions without embodiment mismatch, it enables efficient imitation learning.
By focusing on task-specific end-effector motions, our approach learns skills that transfer to unseen settings, such as new obstacles or changed object positions, from as little as five demonstrations.
We make code and videos available at \website{}.

\end{abstract}
\begin{IEEEkeywords}
Teleoperation, mobile manipulation, imitation learning.
\end{IEEEkeywords}

\section{Introduction}

\IEEEPARstart{W}{hile} robots have reached the hardware capabilities to tackle a wide range of household tasks, generating and executing such motions remains an open problem. The efficient collection of diverse robotic data has become a key factor in teaching such motions via imitation learning~\cite{celemin2022interactive, hussein2017imitation, ravichandar2020, fu2024mobile, honerkamp2024language}.
Although a wide variety of interfaces, teleoperation methods, and kinesthetic teaching approaches exist for static manipulators, collecting demonstrations for mobile manipulation platforms is still challenging. Their large number of degrees of freedom (DoF) often overwhelm standard input methods such as joysticks and keyboards or lead to a large cognitive load when trying to coordinate all the necessary buttons and joysticks. While motion tracking systems~\cite{arduengo2021human, stanton2012teleoperation, krebs2021kit, dass2024telemoma} and exoskeletons~\cite{fu2024mobile, yang2023moma, fang2023low} provide more intuitive interfaces, they are confronted with the correspondence problem if the morphology of robot and human do not match. Furthermore, exoskeletons are highly specialized, expensive equipment, and tracking-based methods restrict the operator from staying within the tracked area, not allowing them to move freely with the mobile robot and having to operate from afar.

\setlength{\tabcolsep}{1pt}
\begin{figure}[t]
	\centering
    \footnotesize
 \includegraphics[width=.9\linewidth,trim={0cm 0cm 0cm 0cm},clip,angle =0,valign=c]{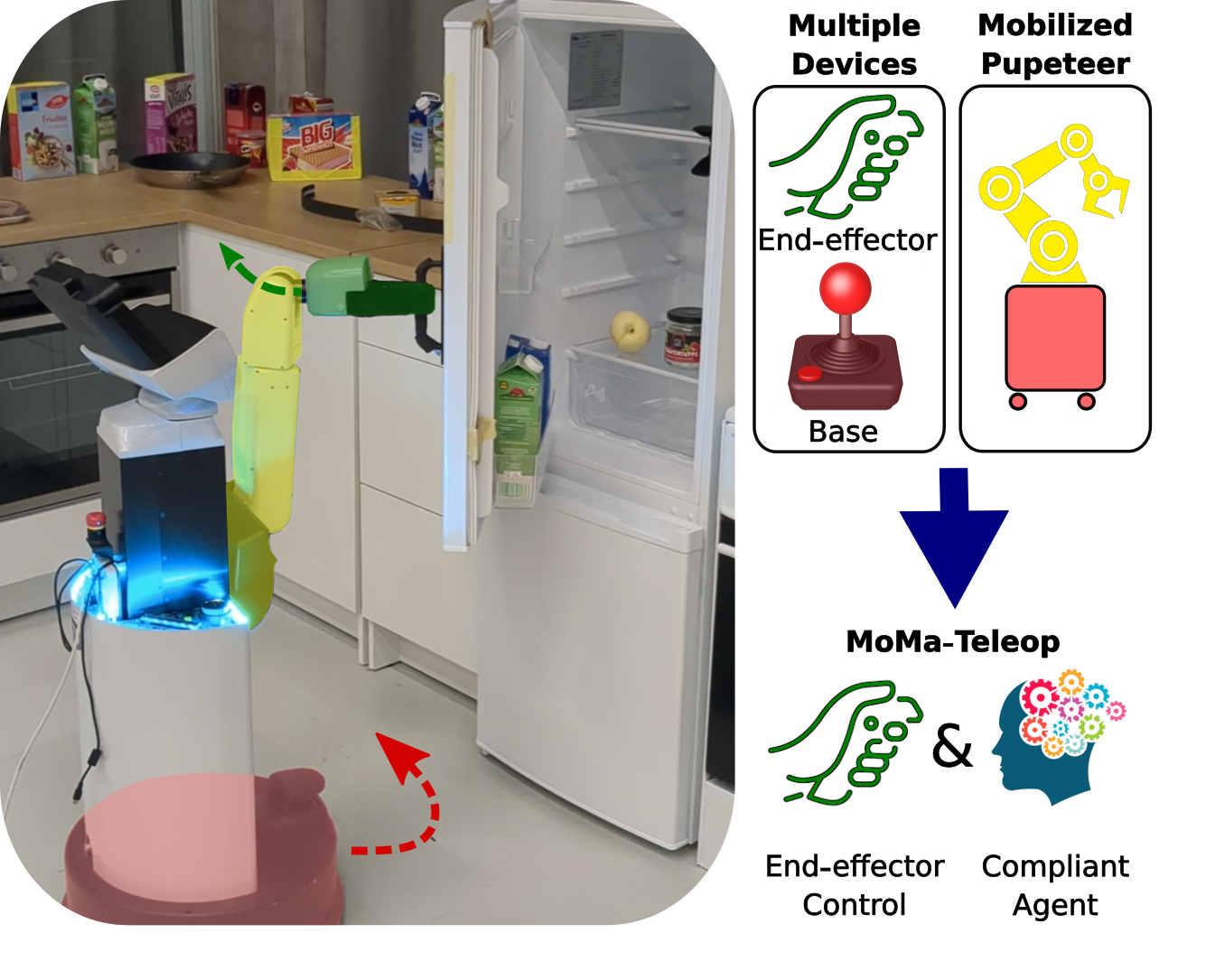}
	\vspace{-0.35cm}
     \caption{Operating mobile manipulators requires to control a large number of degrees of freedom to move {\color{red}{base} (red)}, {\color{Yellow}{arm} (yellow)} and {\color{dark-green}{end-effector} (green)}, requiring multiple input devices or expensive exoskeletons. \ourName{} infers end-effector motions from the operator and communicates them to a reinforcement learning agent to move the base in compliance by converting them to whole-body motions.} 
  	\label{fig:teaser}
\vspace{-0.5cm}
\end{figure}
\setlength{\tabcolsep}{6pt}

We present \ourName{}, shown in \cref{fig:teaser}, an approach for human operation of mobile manipulators that is robot agnostic and requires no additional \myworries{runtime} setup \myworries{or hardware}. We reduce the robot operation problem to pure end-effector motions by delegating the generation of base motions and joint velocities to a \myworries{previously developed \ntwo reinforcement learning agent~\cite{honerkamp2021learning, honerkamp2023learning}}. The operator is solely tasked with controlling the pose of the end-effector through any interface that can generate 6D signals, which we provide for readily available modalities, such as standard joysticks or hand guidance of the end-effector. We then transform this signal into a short-term motion plan for the end-effector that serves as input to the learned base agent. \myworries{This base} agent is trained to control the base while ensuring that the end-effector motions remain kinematically feasible~\cite{honerkamp2021learning, honerkamp2023learning}. It observes the inferred end-effector motions, and can thereby position itself anticipatory with respect to the longer-term intentions of the operator. 

We find that \myworries{our approach} is robust to noisy motion signals generated by humans, as well as fast changes in plans. We demonstrate the resulting capabilities across a wide range of tasks and multiple robots in the real world. Compared to existing teleoperation models, we find that \ourName{} enables significantly faster task completion by generating continuous whole-body motions and alleviating the operator from reasoning about the robot's base. The base agent's dynamic obstacle avoidance enables safe operation via kinesthetic teaching, proactively avoiding collisions with the human teacher and other obstacles in the scene. In contrast to other interfaces, kinesthetic teaching enables the collection of high-quality contact force data, as the operator can physically modulate the desired amount of pressure or force\myworries{~\cite{zhao2022hybrid, yang2023moma}}.

In the second step, we demonstrate that the data generated using our method facilitates efficient learning of the demonstrated tasks. On one hand, the data avoids the correspondence problem between embodiments and is guaranteed to match the robot's kinematic capabilities. On the other hand, the whole-body execution generates smooth motions without base repositioning and re-grasping if a task exceeds the robot's static workspace.
We then show that, by learning task-parameterized end-effector motions~\cite{vonhartz2024imitation} and reusing the learned base agent, our approach can generalize to unseen settings such as new obstacles or object positions from as little as five demonstrations. In contrast, the directly fitted whole-body motions would require much more data to cover these novel scenarios.

To summarize, the main contributions of this work are:
\begin{itemize}
    \item A novel whole-body teleoperation approach for mobile manipulators with a wide variety of input modes.
    \item Mobile manipulation with zero additional setup costs, no workspace restrictions on the operator, and the ability to collect force data via kinesthetic teaching.
    \item Task-centric imitation learning by reusing the same base agent to generalize to new settings.
    \item Extensive real-world experiments across robots and tasks, showing the benefits for both novel and expert users.
    \item We make the code publicly available at \website{}.
\end{itemize}

\begin{table*}[ht]
    \centering
    \caption{Overview of existing Teleoperation Approaches for Mobile Manipulation.} 
    \vspace{-0.5em}
    \begin{threeparttable}
    \renewcommand{\arraystretch}{0.2}
    \setlength{\tabcolsep}{1pt}
    \begin{tabularx}{\textwidth}{lccccccccc}
    \toprule
      & \multirow{2}{*}{\textbf{Cost}} & \multirow{2}{*}{\textbf{Modality}} & \multirow{2}{*}{\textbf{Work Space}} & \multirow{2}{*}{\textbf{Action Space}} & \textbf{Whole-Body} & \textbf{Height}  & \textbf{Robot}   & \textbf{Wrench} & \textbf{Obstacle} \\
      &      &          &           &              & \textbf{Teleop}     & \textbf{Control} & \textbf{Agnostic} & \textbf{Data} & \textbf{Avoidance} \\
      \midrule
      Arduengo~et~al.~\cite{arduengo2021human}         & \textbf{\$\$\$} & Mocap & Tracked Space & EE Pose / Base Vel. & \yes & \yes & \yes & \no & M\\
      MoMaRT~\cite{wong2021erroraware}                 & \textbf{\$} & Phone & Unlimited & EE Pose / Base Vel. & \no & \no & \yes & \no & M\\
      MOMA-Force~\cite{yang2023moma}                   & \textbf{\$\$\$\$} & Kinesthetic & Unlimited & EE Pose and Wrench & \yes & \no & \no & \yes & \no \\
      SATYRR~\cite{purushottam2023dynamic}             & \textbf{\$\$\$\$} & Puppeteer & Unlimited & Joint Pos. / Base Vel. & \yes & \no & \no & \yes & M\\
      TRILL~\cite{seo2023deep}                         & \textbf{\$\$} & VR & Tracked Space & EE Pose / Gait & \yes & \no & \yes & \no & M\\
      Zhao~et~al.~\cite{zhao2022hybrid}                & \textbf{\$\$} & Kinesthetic & Unlimited & EE-Pose / Loco-manip. mode & \yes &  \no & \no & \yes & \no \\
      Mobile ALOHA~\cite{fu2024mobile}                 & \textbf{\$\$\$\$} & Puppeteer & Unlimited & Joint Pos. / Base Vel. & \yes & \no & \no & \no & M\\
      OpenTeach~\cite{iyer2024open}                    & \textbf{\$\$} & VR & Unlimited & Base Translation / EE-Orientation & \no & \yes & \yes & \no & M\\
      Dobb·E~\cite{shafiullah2023bringing}             & \textbf{\$\$} & Puppeteer & Unlimited & EE Pose & \yes & \no & \no & \no & \no\\
      TeleMoMa~\cite{dass2024telemoma}                 & \textbf{[\$, \$\$]} & Multi$^*$ & Unlimited / Tracked Space & EE Pose / Base Vel. / Joint Pos. & \yes & \yes & \yes & \no & M\\
      \textbf{\ourName{}} & \textbf{\$}                 & Multi$^\dagger$ & Unlimited & EE Pose & \yes & \yes & \yes & \yes & A\\
    \bottomrule
    \end{tabularx}
      \begin{tablenotes}[para,flushleft]
       \footnotesize      
       Multi$^*$: Joystick, Spacemouse, Keyboard, RGBD, VR;
       Multi$^\dagger$: Joystick, Kinesthetic, extendable to arbitrary 6-DoF inputs such as VR;
       M: Manual, A: Autonomous. Categories defined in the supplementary.
     \end{tablenotes}
   \end{threeparttable}
    \label{tab:related_work}
    \renewcommand{\arraystretch}{1.0}
\vspace{-0.5cm}
\end{table*}
\section{Related Work}
Teleoperation for mobile manipulation faces the difficulty of operating a large number of degrees of freedom. As a result, large data collection efforts have separated base and arm navigation~\cite{collaboration2023open, brohan2023rt, chebotar2023q, bejczy2020mixed, wong2021erroraware, iyer2024open}, thereby unable to solve more complex mobile manipulation tasks that require base and arm coordination.
While mouse and keyboard~\cite{ratner2015web, dass2024telemoma} or mobile phones~\cite{wong2021erroraware} can be used to send commands, generating coordinated motions for all degrees of freedom simultaneously becomes highly challenging.
Joystick teleoperation configurations commonly use shoulder buttons to switch between control modes, overcrowding the functionality of the buttons, as shown in the supplementary. %

Exoskeletons~\cite{fu2024mobile, yang2023moma, fang2023low, purushottam2023dynamic} or sophisticated avatars~\cite{lenz2023bimanual, schwarz2023robust} can reduce the embodiment correspondence problem, by constraining the human motions. However, they are expensive, robot-specific, and cannot make use of robot kinematics that exceed human motions.
While motion capture systems~\cite{arduengo2021human, stanton2012teleoperation, krebs2021kit} have been successfully used to map human motions to whole-body motions for mobile manipulators, this requires specialized hardware and leads to a correspondence problem if the morphology of robot and human do not match. While keypoint tracking from RGB-D data can alleviate the need for expensive hardware~\cite{dass2024telemoma}, it still has to deal with the correspondence problem and may suffer from less precise estimation.
At the same time, the operator must remain in the tracked workspace and cannot lead the robot over spatially extended tasks.
VR interfaces~\cite{dass2024telemoma, seo2023deep, penco2024mixed, garcia2018robotrix, kazhoyan2020learning}, especially with integrated joystick functionalities, offer enough flexibility for simultaneous base and end-effector commands without overwhelming the user. However, they still require hardware, a camera, and a tracking setup. Approaches without external tracking setup still resort to restricting commands such as only allowing base translation but not end-effector translation~\cite{iyer2024open}.

A number of approaches do not track full motions but infer specific function parameters~\cite{kazhoyan2020learning}, use predefined gait sequences~\cite{seo2023deep}, or match to movement primitives~\cite{penco2024mixed}. In contrast, we fully delegate the control of the remaining body parts to a reinforcement learning agent, such that the user only has to focus on the end-effector. While a number of previous approaches have focused on pure end-effector poses, they collect them with human-carried end-effectors~\cite{shafiullah2023bringing, yang2023moma}, making it infeasible to collect further robot states or to actually teleoperate a full robot.

Kinesthetic teaching, in which the passive joints of the robot are physically moved by humans, avoids any embodiment mismatches. While it is very efficient for static manipulation~\cite{ravichandar2020, calinon2019}, it is not feasible to physically move full mobile robot platforms. Though disturbance observer models have been used to enable some degree of compliance on humanoid robots~\cite{ott2013kinesthetic}.  For mobile manipulators, Zhao~\textit{et~al.}~\cite{zhao2022hybrid} combine a whole-body impedance controller with kinesthetic teaching for the end-effector, with adaptive stiffness for locomotion and manipulation modes. Xing~\textit{et~al.}~\cite{xing2021human} use admittance and nullspace control to teach carrying heavy objects. However, as these controllers have no awareness of their environment, they are unable to avoid collisions. In contrast, our approach is able to avoid collisions and position itself anticipatory for the continuation of the end-effector motions. We provide an overview of existing teleoperation systems for mobile manipulation in comparison to our approach in \cref{tab:related_work}. %

\section{\ourName{}}\label{sec:approach}

We aim to reduce the complexity of operating mobile manipulators via existing, standard interfaces. To do so, we develop three components: a user interface that takes in \mbox{6-DoF} signals, an inference module that translates these signals into end-effector motions and a base agent that moves the robot's base to support the desired end-effector motions. An overview of the proposed approach is depicted in \cref{fig:overview}. The result is a modular whole-body teleoperation system that reduces the complexity for the operator to pure end-effector control.

\subsection{Background: Learning Feasible Base Motions}
\begin{figure*}[t]
	\centering
	\resizebox{.73\linewidth}{!}{%
 \includegraphics[width=\linewidth,trim={0cm 0cm 0cm 0cm},clip,angle =0,valign=c]{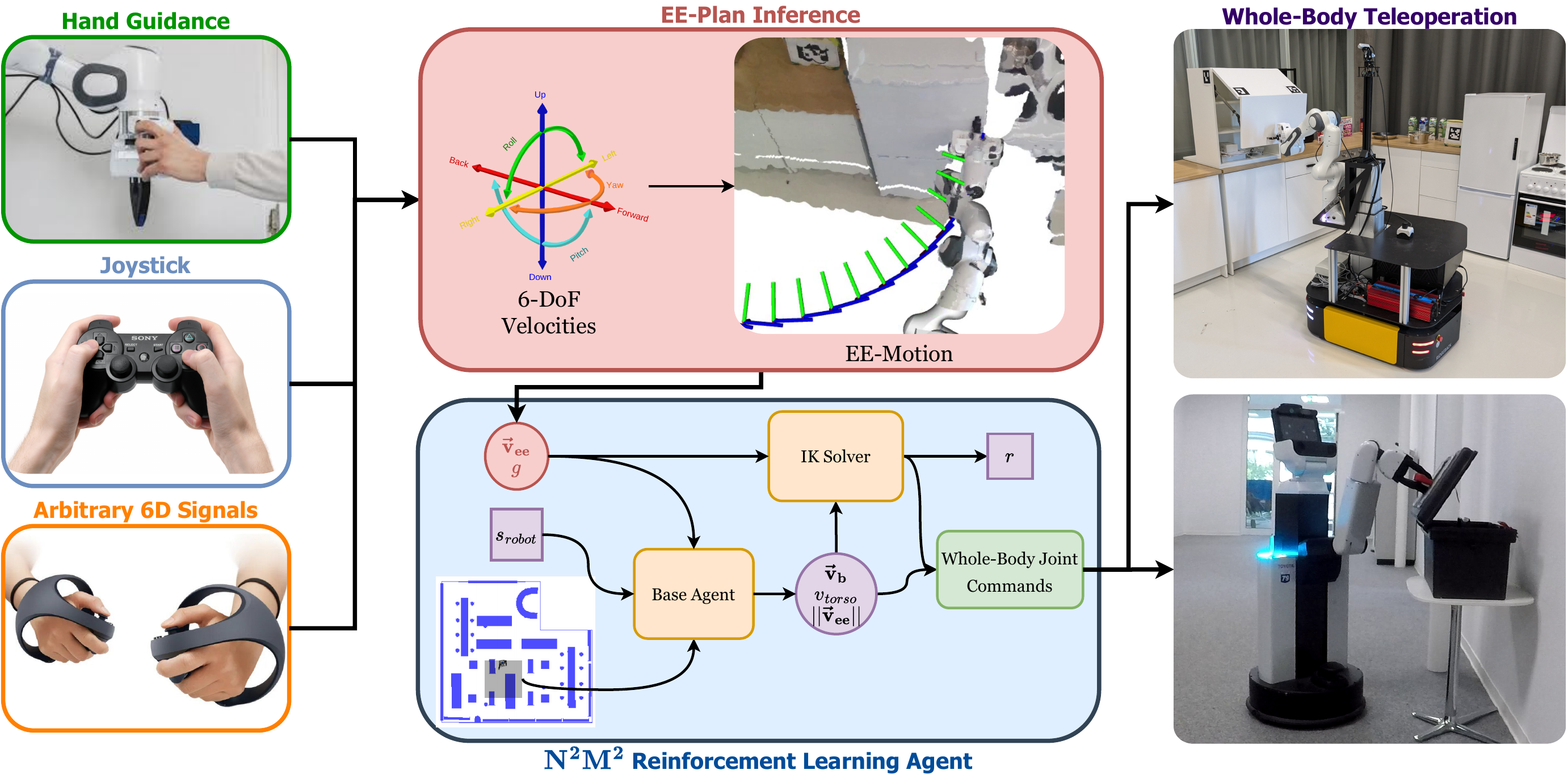}}
     \caption{\ourName{}: We modularize teleoperation for mobile manipulators. The human operator controls the end-effector of the robot, through a range of possible interfaces. A reinforcement learning agent then transforms these commands into whole-body commands, moving the base in compliance to achieve the operator's desired motions, while considering the robot's kinematics and obstacle constraints. \looseness=-1} 
  	\label{fig:overview}
\vspace{-0.5cm}
\end{figure*}

We use our previously developed \ntwo{} approach to decouple end-effector motions from the remaining joint motions and delegate these motions to a reinforcement learning agent for the base of the robot~\cite{honerkamp2021learning, honerkamp2023learning}. This agent, shown in \myworries{the supplementary material and} the blue part of \cref{fig:overview}\myworries{, is tasked to convert end-effector motions to whole-body motions. It }receives a desired end-effector motion, consisting of translation and orientation velocities $\vec{v}_{ee}$ to the next desired pose as well as a more distant end-effector subgoal $g$ in the form of a 6-DoF pose in the base frame of the robot that indicates the longer-term plan. 
The agent then generates velocities $\vec{v}_b$, $v_{torso}$ for the base and torso of the robot and uses inverse kinematics for the remaining arm joints, thereby completing the whole-body motions. It also learns to regularize the speed at which the end-effector motions are executed through a scaling factor $||\vec{v}_{ee}||$. Based on a local occupancy map, it learns to avoid obstacles. At test time, the agent generalizes to unseen end-effector motions and can dynamically react to static and dynamic obstacles, which was demonstrated in a number of works using these policies~\cite{schmalstieg2023learning, honerkamp2024language}. We leverage this ability to enable arbitrary, unseen end-effector motions in this work.

\subsection{Interfaces}
Given the base agent, the human operator is tasked with generating 6-DoF velocities for the end-effector. We implement methods for a range of common, low-cost interfaces, including joysticks and hand guidance.
However, our approach is compatible with any modality that can generate such a signal. Importantly, these interfaces are already existing or extremely \myworries{affordable} mobile interfaces, without any workspace restrictions of cameras or tracking systems for the operator.\looseness=-1

\setlength{\tabcolsep}{1pt}
\setlength{\tabcolsep}{6pt}

\para{Joystick:} 
As we reduce the required inputs to 6-DoF for the end-effector, we can comfortably serve all inputs simultaneously on a standard Dualshock3 joystick, shown in the supplementary. We use the left and right controller sticks together with two shoulder buttons for translation and orientation changes. These commands are applied in the frame of the wrist camera that is streamed to the user. Two additional buttons enable the opening and closing of the gripper. Lastly, we add a button to switch to a higher-precision mode with smaller end-effector velocities, as discussed in \cref{sec:eemotions} below. This results in a reduction from 18 buttons in default teleoperation down to 10 buttons that can all be operated simultaneously.\looseness=-1

\para{Hand Guidance:} In this mode, a human physically guides the manipulator arm to kinesthetically teach the robot. The physical guidance has the particular benefit of being able to demonstrate specific wrenches for contact-rich tasks\myworries{, which was shown to provide valuable signals for imitation learning policies\cite{yang2023moma, zhao2022hybrid, liu2024forcemimic}}. With our approach, we are able to extend this method from static arms to mobile manipulators. The operator moves the end-effector of the robot, and we \myworries{track} changes in translation and orientation of the end-effector as motion signals for the base agent. The gripper can be opened and closed via a button on the end-effector.
To ensure safety, a deadman switch on the end-effector immediately stops the robot if it is released. As this requires the human operator to move next to the robot, avoiding collisions is essential for safe operation. The base agent detects the human as an obstacle in its LiDAR scan and reacts immediately to the human's movements. We verify this capability in our experiments. The agent's velocity scaling is ignored in this setting.

\para{Arbitrary input modalities:} While we provide implementations for joystick and hand guidance, our approach can be used with any input modality that can generate 6-DoF input signals, such as VR devices or SpaceMouses.

\subsection{Inferring End-Effector Motions}\label{sec:eemotions}
After getting pose and velocity signals from the teleoperation interface, we transform these measurements into end-effector translation and orientation deltas consisting of a 3D velocity vector $v_{\mathit{signal}}$ and the change in orientation converted to a unit quaternion $q_{\mathit{signal}}$. To enable the reinforcement learning agent to make good decisions, we extrapolate these deltas into an end-effector motion $m_{ee}$. This motion consists of a vector of 6D end-effector poses, spaced at a fixed resolution of $res_{\mathit{training}} = \SI{0.1}{\meter}$ over a distance of up to $d_g = \SI{1.5}{\meter}$ into the future. From this motion, the agent infers the next desired end-effector velocities and the last pose as a subgoal.
We update the inferred end-effector motions at high frequency from the latest user inputs, enabling quick reaction to changes\footnote{We run the complete system at around \SI{30}{\hertz} on the compute limited HSR and at around \SI{80}{\hertz} on the FMM robot.}.

\subsubsection{User Signal} In the following we describe how we infer these directional and translation signals $v_{\mathit{signal}}$ and $q_{\mathit{signal}}$ from different interfaces.

\para{Joystick:} We directly map the pressed joystick buttons to translation and rotation commands based on the button assignment shown in the supplementary.
As the control over speed is delegated to the RL agent and it can change the norm of the vector to this next desired pose by a factor $[0.01, 2]$, we normalize the translational velocity vector to match the resolution used during training of the RL agent, $||v_{\mathit{signal}}|| = res_{\mathit{training}}$ and scale the angular velocities into a range of $[0, 0.1875]\, \si{\radian}$. 
We allow the user to switch to a high-precision mode via the press of a button. In this mode, we only allow the RL agent to slow down the velocities by clipping the learned velocity scaling action at 1.0. In addition, we reduce the horizon of the end-effector plan (cf. functional form below) to $d_g = \SI{0.3}{\meter}$. %

\para{Hand Guidance:} We infer the signal from the operator's movement of the end-effector. We record a history of the robot's end-effector poses $ee_t$, consisting of a tuple of position and orientation $(ee^{\mathit{pos}}_{, t}, ee^q_{t})$, over a time of $h=\SI{1}{\sec}$ at a rate of \SI{33}{\hertz}. If the user pauses and stops sending signals, we reset the history. As physical guidance motions can be noisy, we first smoothen the input signals: we calculate $v_{\mathit{signal}}$ as a weighted sum of the first differences with exponential weighting, where $H$ is the number of end-effector poses in the history:
\begin{equation}\label{eq:weights}
    v_{\mathit{signal}} = \sum_{h=0}^{H-1}\frac{1}{2^h} (ee^{\mathit{pos}}_{t-h} - ee^{\mathit{pos}}_{t-h-1}).
\end{equation}
We then assume a maximum translational velocity generated by the user of $v^{\mathit{trans}}_{\mathit{max}} = \SI{0.125}{\meter\per\second}$ and re-normalize the observed signals to a range of $[0, d_g]$.
For orientations, we similarly calculate a weighted average of the changes in orientation $q_{\mathit{delta}} = ee_{q, t} \cdot ee_{q, t-1}^{-1}$. To do so, we average the corresponding quaternions with the same weights as above through minimization of the attitude matrix differences~\cite{markley2007averaging}. We then apply this change in rotation to the current end-effector orientation $q_{\mathit{signal}} = \avg(q_{\mathit{delta}})^n$ where we empirically set $n=3$.
\myworries{As these signals rely on simple computations, we can also easily infer them at higher frequencies beyond \SI{100}{\Hz}, ensuring small differences in poses, such that they remain accurate approximations even for faster motions.}

\subsubsection{Functional form}
Next, we transform the user signals to a longer end-effector motion to communicate the user intentions to the base agent. The end-effector motions that the robot has to execute locally follow linear motions and smooth curves. To achieve this, we chose a form of linear dynamic system to extrapolate the inferred signals, enabling the execution of arbitrary motions.
In particular, we extrapolate the velocities by integrating a dynamic linear system to infer the end-effector motion $m_{ee} = [ee_{t+i} | i \in 1, ..., T]$:
\begin{equation}\label{eq:dynamic_system}
    ee_{t+1} = (ee^{\mathit{pos}}_{t} + q_{\mathit{signal}} \cdot v_{\mathit{signal}},\,q_{\mathit{signal}} \cdot ee^{q}_{t}).
\end{equation}
We run this system for $T = \max(||v_{\mathit{signal}}|| \cdot \frac{d_g}{res_{\mathit{training}}}, 5)$ steps, thereby matching the planning horizon $d_g = \SI{1.5}{\meter}$ used during training and producing shorter plans for small (unnormalized) velocity signals $||v_{\mathit{signal}}||$.
The resulting motions are shown in the supplementary and the video.

\subsection{RL Base Agent}
The resulting motions $m_{ee}$ are then provided to a pretrained N$^2$M$^2$ base agent, where we replace the end-effector motion module used during training with the one above. The agent is active whenever the human operator generates a signal and immediately pauses whenever the signal stops (all joystick buttons or hand guidance deadman switch released).
\myworries{Pretrained checkpoints are publicly available~\cite{honerkamp2023learning}. Adaptation to new robots encompasses a training cost of around \SI{4}{\hour}.}
\myworries{Note that our approach is agnostic to the exact form of this component. On robots with lower degrees of freedom, it could also be coupled with inverse kinematics based works~\cite{takeshita2024whole}.}\looseness=-1

\begin{figure*}
    \centering
    \captionsetup[subfigure]{justification=centering}
    \subfloat[\texttt{Pick \& Place}\\ \vspace{0.14em}\footnotesize{Pick bottle from coffee table (purple) and place it on a high shelve (green).\looseness=-1}]{\adjincludegraphics[width=0.24\linewidth,trim={0cm {.175\height} 0cm 0cm},clip]{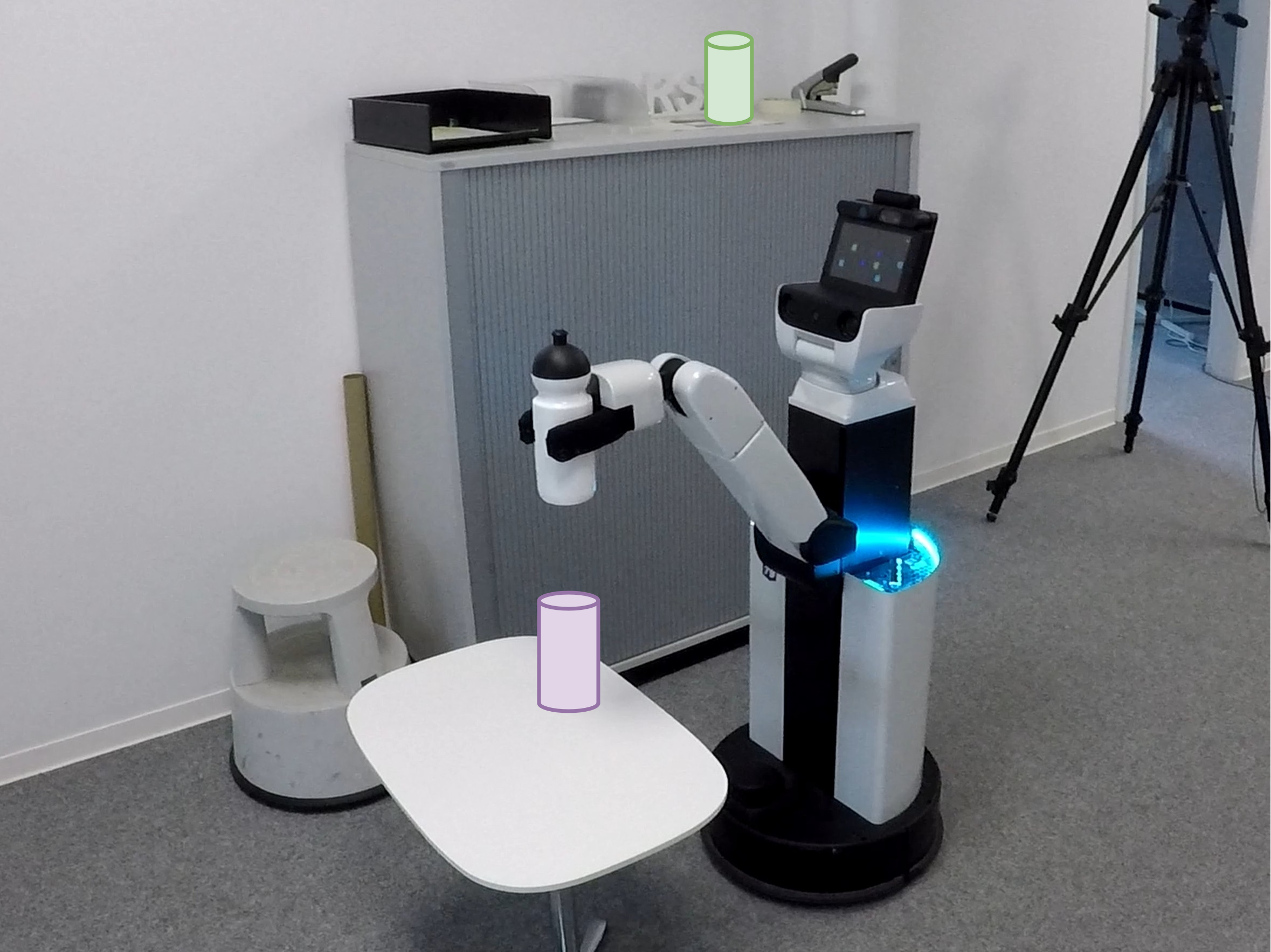}}\hspace{0.25pt}
    \subfloat[\texttt{Microwave}\\ \vspace{0.14em}\footnotesize{Open a microwave in a narrow kitchen.}]{\adjincludegraphics[width=0.24\linewidth,trim={0cm 0.0cm 0cm {.175\height}},clip]{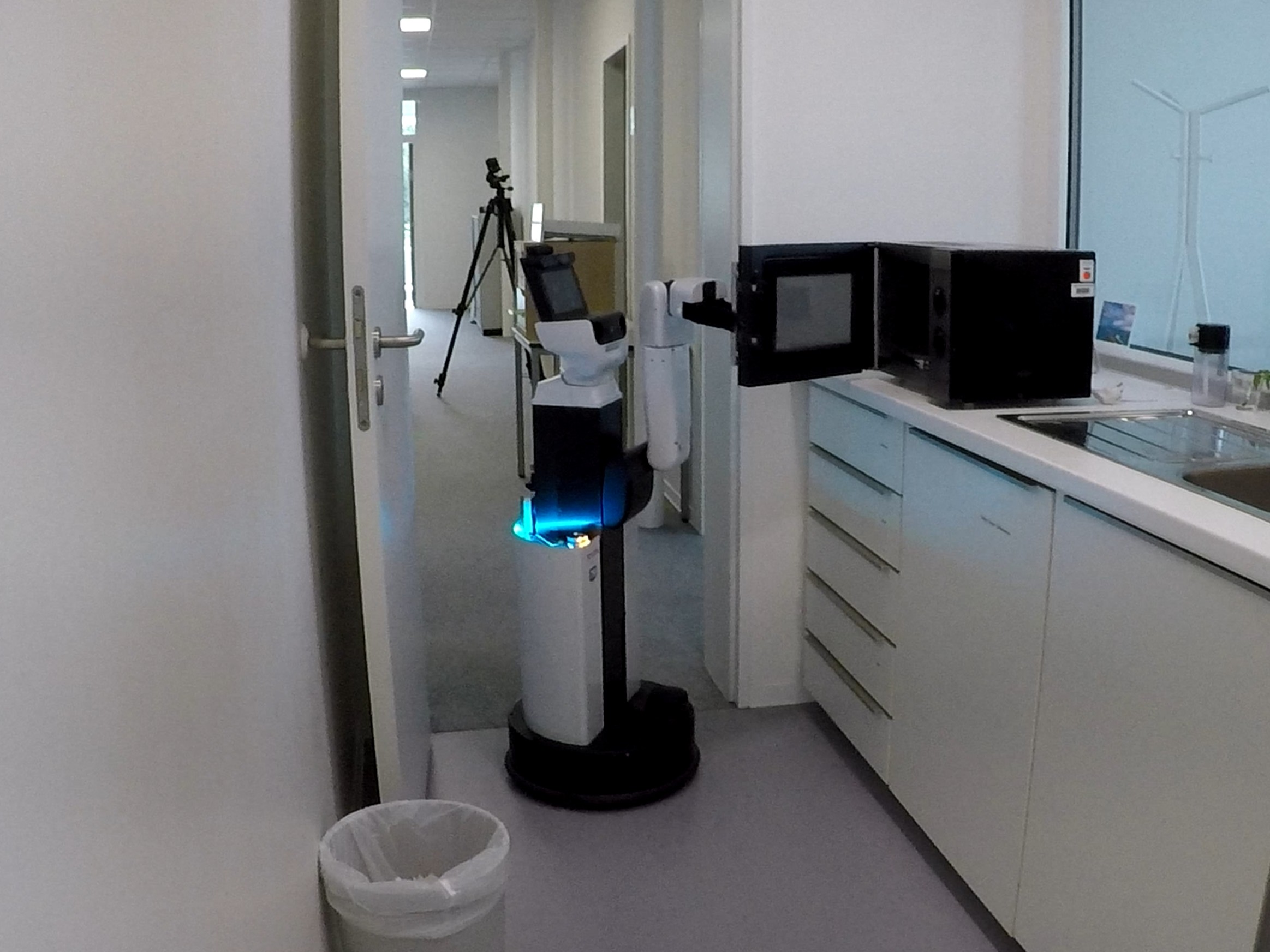}}\hspace{0.25pt}
    \subfloat[\texttt{Clean Table}\\ \vspace{0.14em}\footnotesize{Scrub a predefined pattern with a sponge.}]{\adjincludegraphics[width=0.24\linewidth,trim={0cm {.175\height} 0cm 0cm},clip]{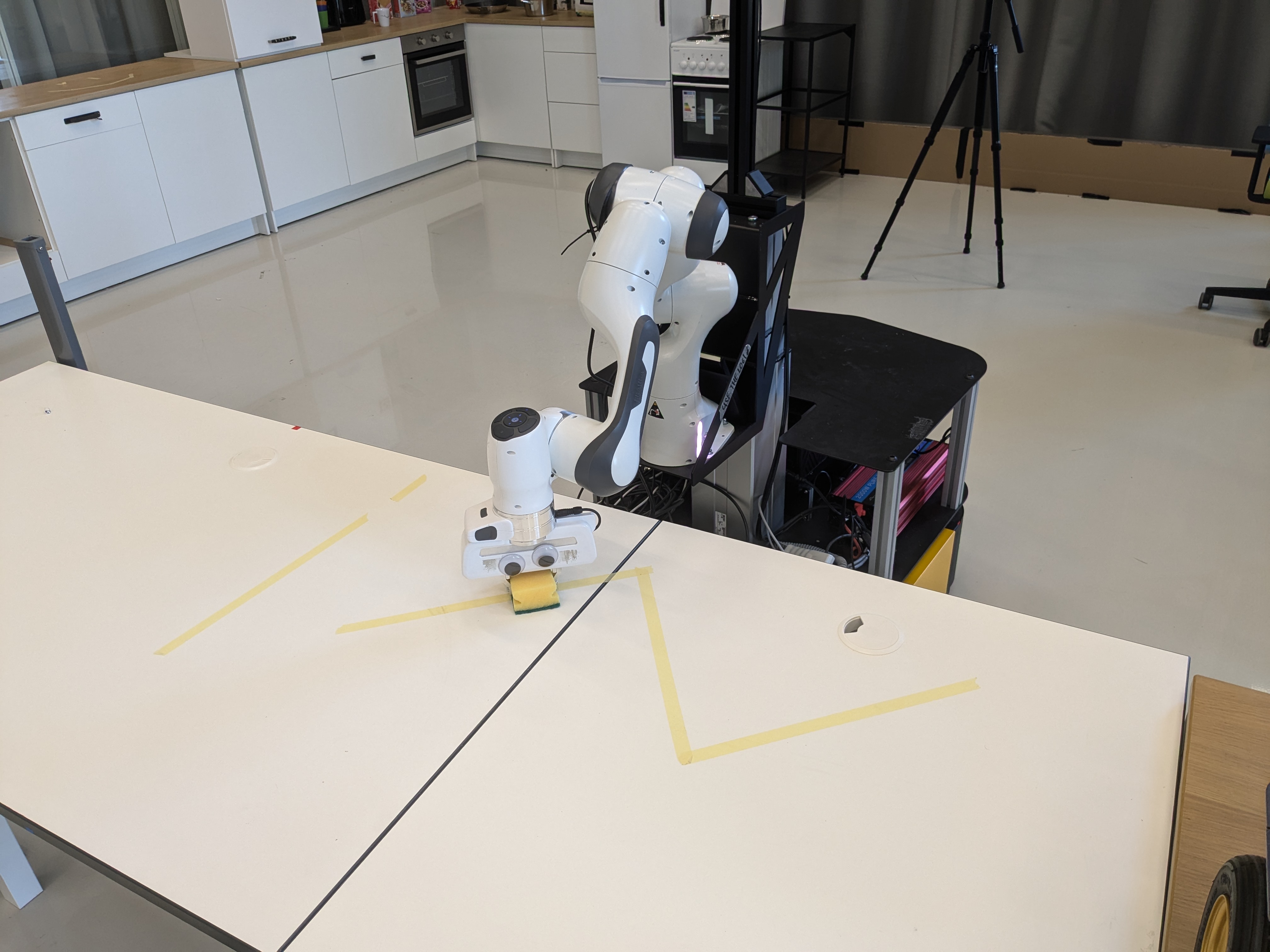}}\hspace{0.25pt}
    \subfloat[\texttt{Door Outwards}\\ \vspace{0.14em}\footnotesize{Open a door outwards.}]{\adjincludegraphics[width=0.24\linewidth,trim={0cm 0.0cm 0cm {.175\height}},clip]{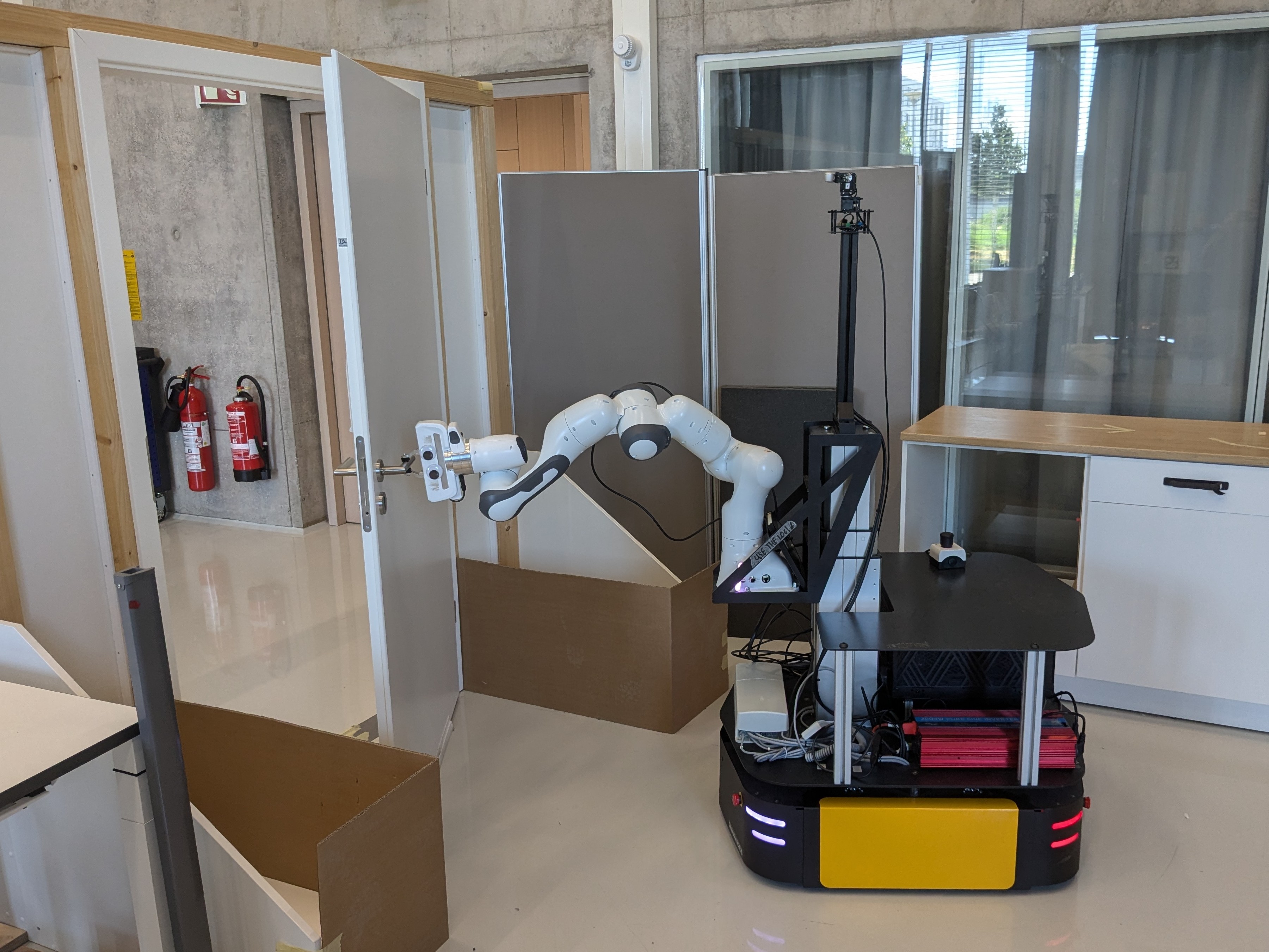}}
    \\ \vspace{-0.15cm}
    \subfloat[\texttt{Door Inwards}\\ \vspace{0.14em}\footnotesize{Open a door inwards, driving through the frame while grasping the handle.}]{\adjincludegraphics[width=0.24\linewidth,trim={0cm 0.0cm 0cm {.175\height}},clip]{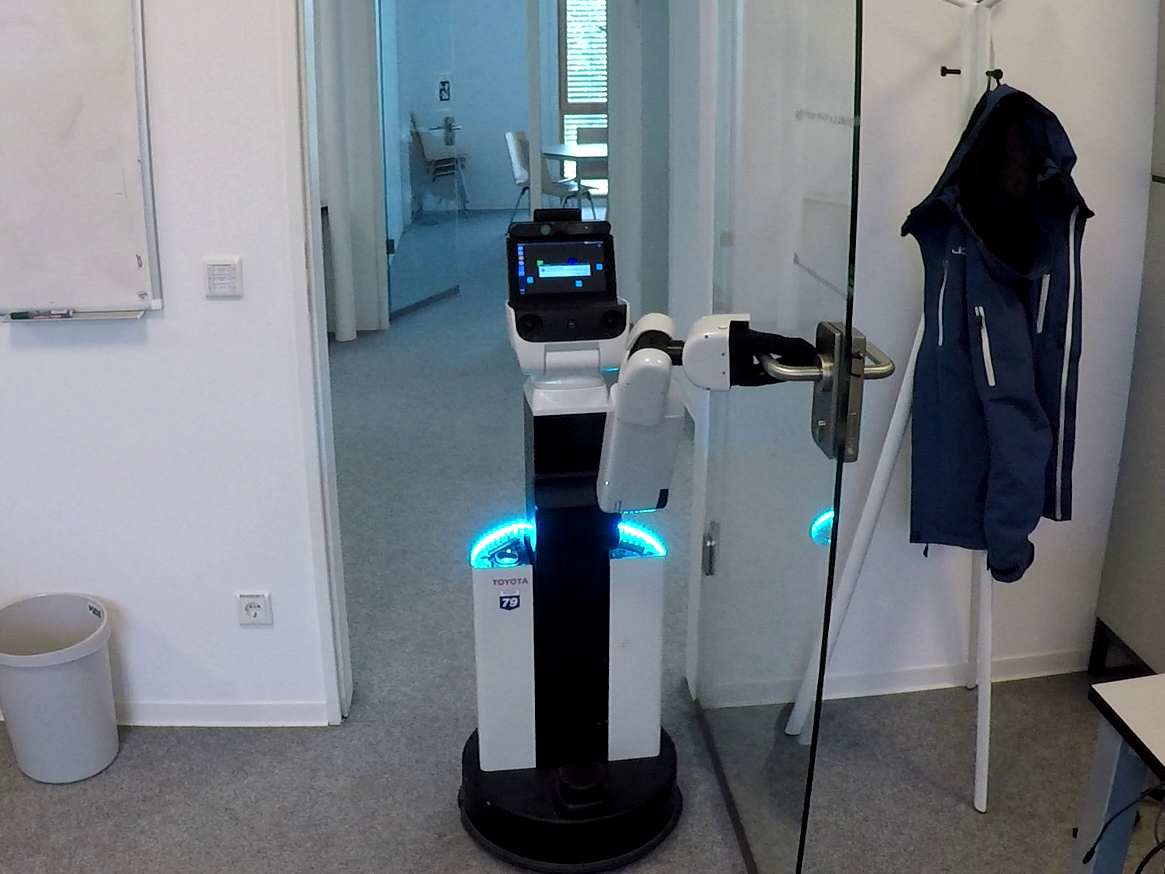}}\hspace{0.25pt}
    \subfloat[\texttt{Toolbox}\\ \vspace{0.14em}\footnotesize{Open a toolbox with an upwards revolute joint.}]{\adjincludegraphics[width=0.24\linewidth,trim={0cm {.175\height} 0cm 0cm},clip]{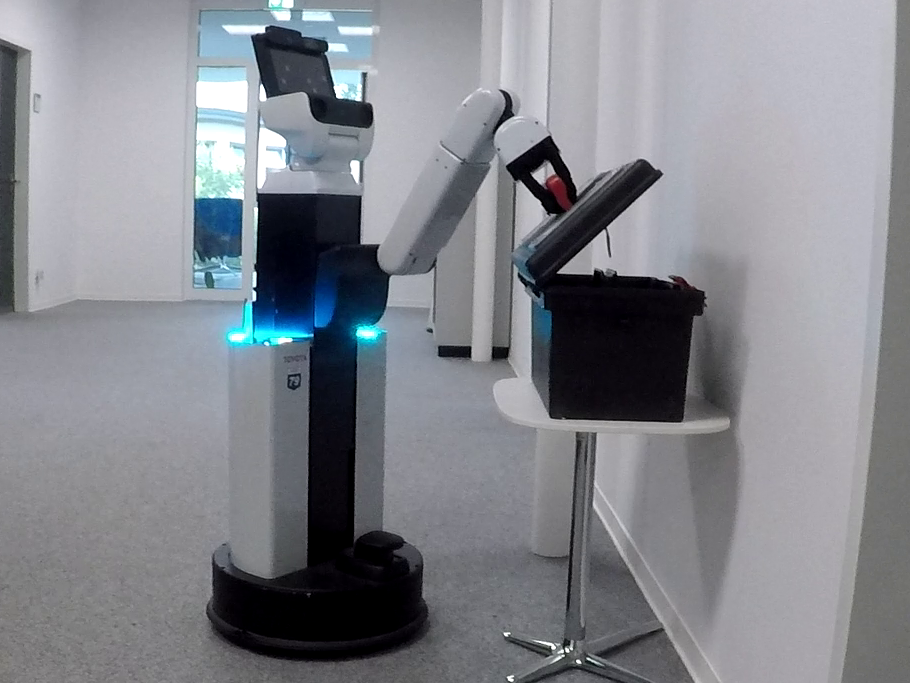}}\hspace{0.25pt}
    \subfloat[\texttt{Folding Cabinet}\\ \vspace{0.14em}\footnotesize{Open a cabinet with a complex folding articulation.}]{\adjincludegraphics[width=0.24\linewidth,trim={0cm {.175\height} 0cm 0cm},clip]{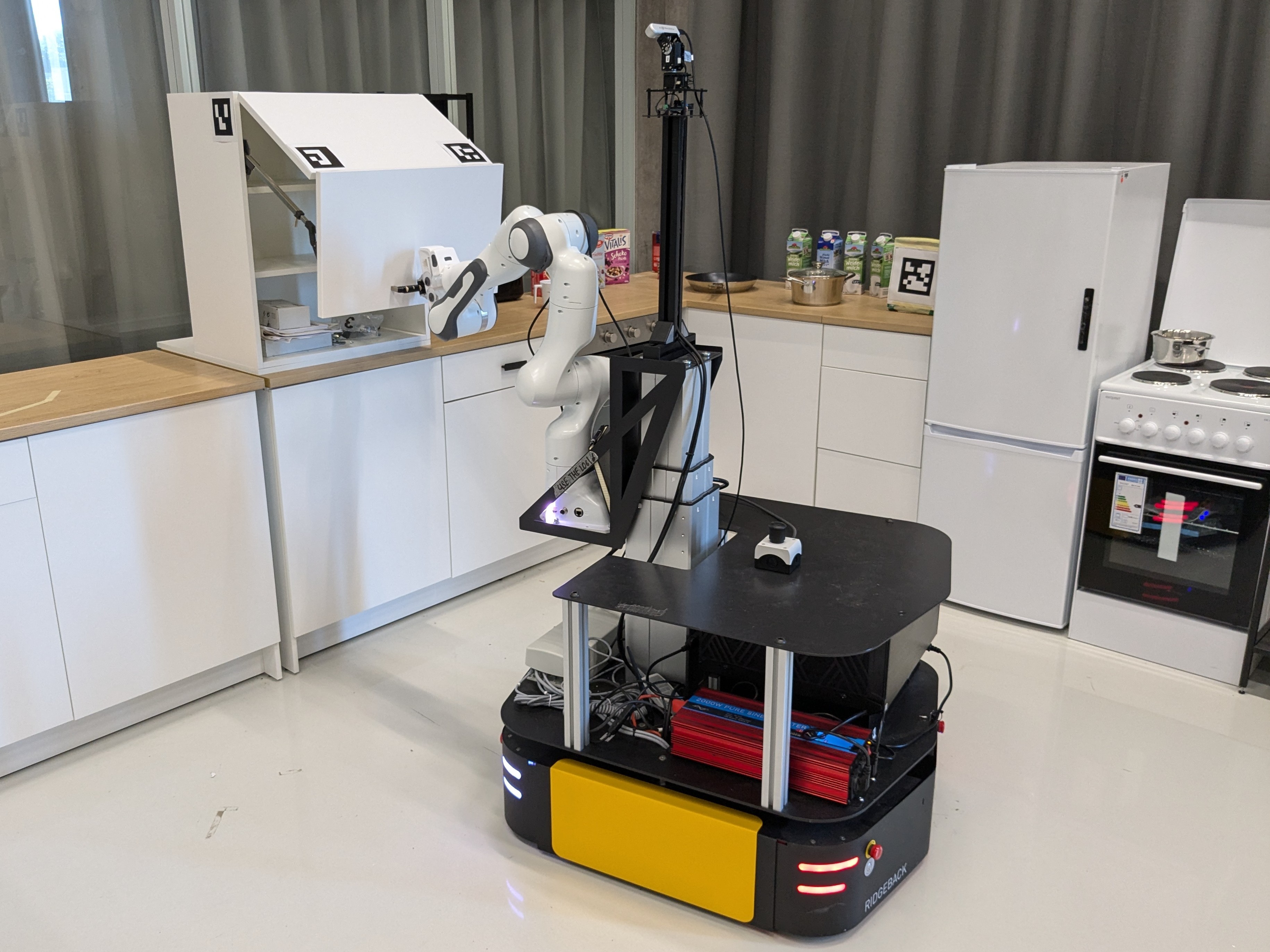}}\hspace{0.25pt}
    \subfloat[\texttt{Fridge Pick \& Place}\\ \vspace{0.14em}\footnotesize{Open the fridge, pick a milk box and place on the shelve next to the oven.}]{\adjincludegraphics[width=0.24\linewidth,trim={0cm {.175\height} 0cm 0cm},clip]{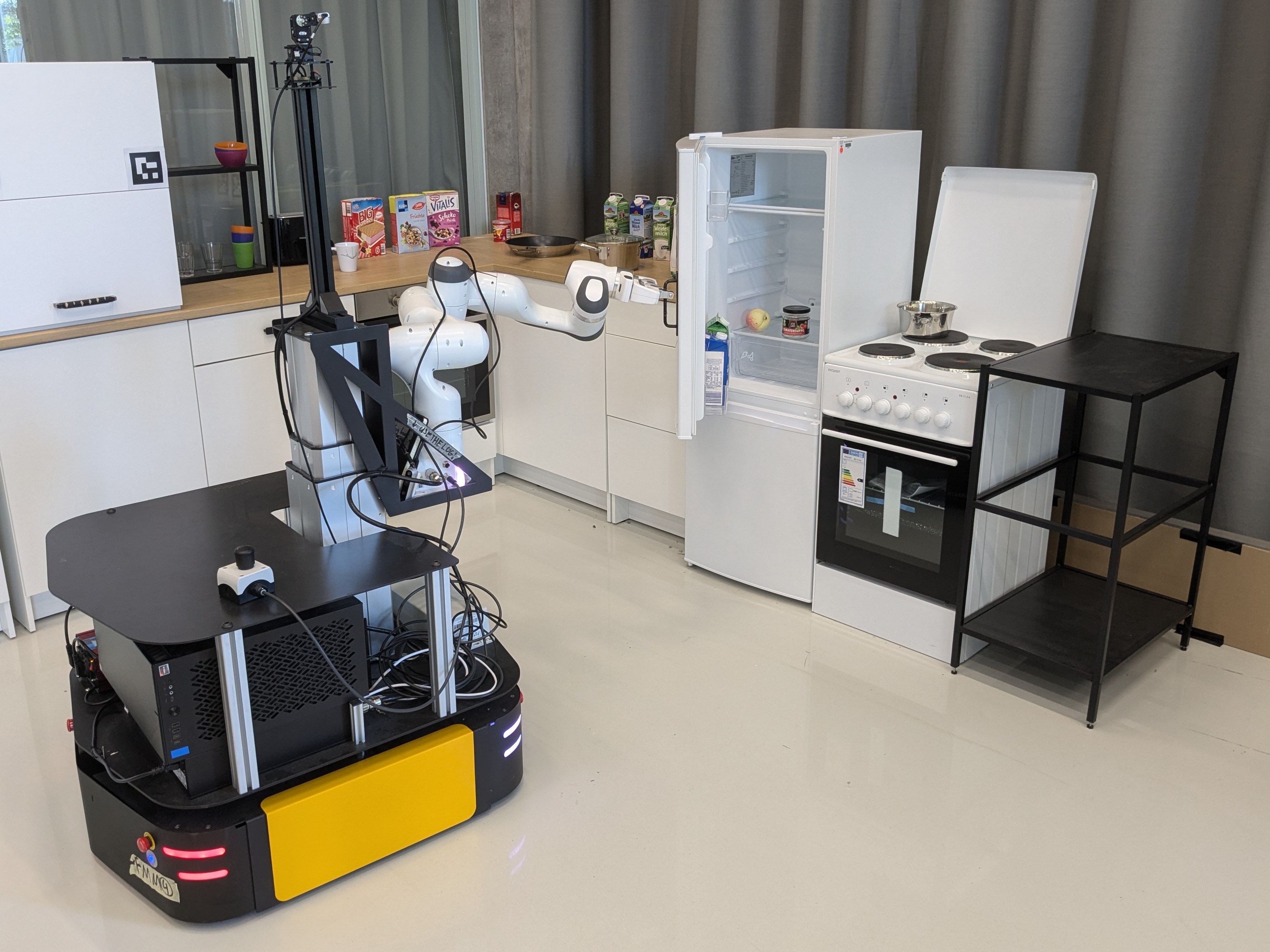}}
     \caption{Teleoperation tasks on the HSR (left) and FMM (right) robots.}\label{fig:tasks}
    \vspace{-0.35cm}
\end{figure*}

\section{Teleoperation Experiments}

\para{Robots:}
The \textit{HSR} robot has an omnidirectional base and a 5-DoF arm, including a torso lift joint, resulting in 8-DoF. The 11-DoF \textit{FMM} robot consists of an omnidirectional Ridgeback base with a lifting column and a Franka 7-DoF Arm.

\para{Tasks:} We evaluate the approaches on a wide range of tasks that cover a diverse set of motions, contact-rich manipulation as well as operation in narrow spaces. 
We start all experiments from a default start configuration and fixed start position. To demonstrate the compatibility with different input modalities, we evaluate the HSR in teleoperation mode and the FMM with kinesthetic teaching via hand guidance. The tasks are described in \cref{fig:tasks} and the supplementary.

\para{Baselines:} We compare our approach against a diverse set of approaches based on different input modalities. We focus on methods with comparably low setup costs to our approach.\\
\para{Joystick:} On the HSR, we use the Dualshock3 joystick teleoperation package that was developed for the robot, shown in the supplementary. It uses buttons to directly send joint commands for the arms. Shoulder buttons serve as switches to change between base and arm commands. As a result, it is not possible to simultaneously send base and arm commands.\\
\para{Hand Guidance:} On the FMM, we use static hand guidance control for the arm of the robot, combined with control of the base and lift joint via the joystick.\\
\para{Vision Tracking~\cite{dass2024telemoma}} tracks human motions with an RGB-D camera and translates torso movements and relative hand movements to robot motions, using inverse kinematics to calculate the connecting arm joints.\\
\para{VR Tracking~\cite{dass2024telemoma}} translates motions from virtual reality handheld controllers to end-effector motions. We implement the approach with HTC Vision Pro with two lighthouse towers. For the FMM, we set the Franka arm to a low stiffness, enabling it to make contact with the objects safely. Higher stiffness resulted in hardware safety violations upon contact.

\para{Metrics:} We compare the average \textit{success rate (SR)} over the attempted tasks. Failures can stem from reaching safety limits or collisions with the environment. Moreover, we measure the average \textit{completion time} for the task, averaged over \textit{successful} executions only.

\subsection{\myworries{Teleoperation} Evaluation}\label{sec:teleop_evaluation}
\begin{table*}
    \centering
    \caption{Teleoperation results across robots and tasks.} 
    \vspace{-0.5em}
    \begin{threeparttable}
    \begin{tabularx}{0.89\textwidth}{llcccccccccc}
    \toprule
      \multicolumn{2}{c}{\textbf{HSR Robot}} & \multicolumn{2}{c}{P\&P} & \multicolumn{2}{c}{Microwave} & \multicolumn{2}{c}{Door Inwards} & \multicolumn{2}{c}{Toolbox} & \multicolumn{2}{c}{Average} \\
      \cmidrule(lr){3-4}
      \cmidrule(lr){5-6}
      \cmidrule(lr){7-8}
      \cmidrule(lr){9-10}
      \cmidrule(lr){11-12}
            Model & Modality          & SR & Time & SR & Time & SR & Time & SR & Time & SR & Time \\
      \midrule
      Joystick & Joystick               & \textbf{100} & 42.0 & \textbf{100} & 42.0 & \textbf{100} & 66.8 & \textbf{100} & 83.6 & \textbf{100} & 58.6\\
      Vision Tracking & Camera          & \phantom{0}40 & 41.0 & \phantom{0}60  & 43.7 & n.e. & n.e. & n.e. & n.e.  & \phantom{0}25 & (42.4)\\
      VR Tracking & VR + Camera         & \textbf{100} & \textbf{38.4}& \phantom{0}80  & 46.5 & 0/(80$^*$) & (84.5$^*$) & \phantom{00}0 & --  & \phantom{0}45 & (42.5) \\
      \ourName{} & Joystick             & \textbf{100} & 44.2 & \textbf{100} & \textbf{36.2} & 80 & \textbf{46.3} & \textbf{100} & \textbf{55.2}  & \phantom{0}95 & \textbf{45.5}\\
    \midrule
    \midrule
      \multicolumn{2}{c}{\textbf{FMM Robot}} & \multicolumn{2}{c}{Clean Table} & \multicolumn{2}{c}{Door Outwards} & \multicolumn{2}{c}{Folding Cabinet} & \multicolumn{2}{c}{Fridge P\&P} & \multicolumn{2}{c}{Average} \\
      \cmidrule(lr){3-4}
      \cmidrule(lr){5-6}
      \cmidrule(lr){7-8}
      \cmidrule(lr){9-10}
      \cmidrule(lr){11-12}
            Model & Modality          & SR & Time & SR & Time & SR & Time & SR & Time & SR & Time \\
            \midrule
    Hand Guidance & Hand Guidance + Joystick  & \textbf{100} & 42.8 & \phantom{0}80 & 77.8 & \textbf{80} & 62.5 & \textbf{100} & 81.2 & \phantom{0}90 & 66.1 \\
    Vision Tracking & Camera                & n.e. & n.e. & n.e. & n.e. & n.e. & n.e. & n.e. & n.e. & n.e. & n.e.\\
    VR Tracking & VR + Camera               & \textbf{100} & 114 & \phantom{00}0 & -- & n.e. & n.e. & n.e. & n.e. & \phantom{0}25 & (114) \\
    \ourName{} & Hand Guidance               & \textbf{100} & \textbf{38.4}& \textbf{100} & \textbf{43.0} & \textbf{80} & \textbf{43.3} & \textbf{100} & \textbf{62.6} & \textbf{\phantom{0}95} & \textbf{46.8}\\
    \bottomrule
    \end{tabularx}
        \begin{tablenotes}[para,flushleft]
       \footnotesize      
       SR: average success rate in percent, time: average completion time in seconds over the \textit{successful} attempts, n.e.: not evaluated on this task due to hardware safety concerns, $^*$: finished opening door, but was unable to grasp and follow the handle of the door.
     \end{tablenotes}
   \end{threeparttable}
    \label{tab:teleop}
\vspace{-0.5cm}
\end{table*}

We execute all methods five times per task, resulting in 20 episodes per robot and method. The evaluations are conducted by an operator with several hours of experience in all methods. The results are reported in \cref{tab:teleop} and shown in the video.
We find that the static operation methods via joystick or hand guidance, as well as our approach, are able to complete all tasks successfully. The tracking approaches are efficient on tasks such as \texttt{Pick \& Place} or \texttt{Microwave} with the fastest completion time in the former. However, we find that the tracking approaches become too imprecise for tasks requiring higher precision or movement over larger distances and rotations. Particular difficulties include the limited operator workspace, operation from the tracked area afar and the embodiment mismatch between operator and robot, see the supplementary for additional details.
As this resulted in frequent safety limit violations and emergency stops of the robots, we abstain from evaluating them on the remaining tasks to ensure the safety of the equipment.

Pure joystick operation is very robust and can efficiently complete tasks such as opening a microwave or door in which we can keep the relative end-effector pose constant and use pure base motion for translation and yaw changes. However, tasks such as opening the toolbox that requires backward movement together with arm translation and pitch changes of the end-effector become tedious, involving numerous switches between base and arm motions.
Similarly, for static hand guidance, tasks such as opening a door become cumbersome, as they require repeated base repositioning. Moving the base while in contact with the door risks triggering safety limits of the Franka arm, requiring to release the door handle, move the base, and then re-grasp.
In contrast, we found \ourName{} to enable continuous operation during these tasks. The human can operate directly next to the robot, and the base agent is compliantly moving the base while considering the robot's kinematics and obstacles. Cleaning the table, we found hand guidance based methods to result in good tracking of the pattern with constant contact of the table. In contrast, with tracking-based methods, it was difficult to keep contact through the full scrubbing motion.
Overall, we found that \ourName{} facilitated substantially faster and more continuous task executions across tasks, robots, and input modalities\myworries{, completing tasks significantly more consistently than the tracking methods and significantly faster than the similarly successful baselines on six out of eight tasks. We report standard errors and significance in the supplementary material.}

\subsection{User Study}
\pgfplotsset{select coords between index/.style 2 args={
    x filter/.code={
        \ifnum\coordindex<#1\def\pgfmathresult{}\fi
        \ifnum\coordindex>#2\def\pgfmathresult{}\fi
    }
}}

\pgfplotsset{%
    width=.60\linewidth,
    height=5.3cm
}

\begin{figure}[t]
\centering
\footnotesize
\begin{tikzpicture}
\begin{axis}[
    ybar,
    axis lines*=left,
    legend style={at={(1,0.5)},
      anchor=west,legend columns=1, draw=none},
    ylabel={Completion Time (\si{\second})},
    symbolic x coords={door,fridge},
    xtick=data,
    nodes near coords,
    nodes near coords align={vertical},
    every node near coord/.append style={yshift=6pt,font=\footnotesize},
    enlarge x limits=0.5,
    ybar=6pt,
    ymin=0,
    ymax=125,
    bar width=12pt,
    xticklabels = {
            Door Outwards,
            Fridge P\&P,
        },
  xticklabel style={
        text width={2.0cm},
        align={center},
      },
      legend style={/tikz/every even column/.append style={column sep=0.1cm}}]
    ]
\addplot[black,fill=blue!60] plot[error bars/.cd, y dir=both,y explicit] coordinates {(door,115.5) +- (9.09,9.09) (fridge,79.6) +- (6.33,6.33)};
\addplot[black,fill=orange!100] plot[error bars/.cd, y dir=both,y explicit] coordinates {(door,55.7) +- (4.87,4.87) (fridge,70.9)  +- (6.31,6.31)};
\addplot+[black,mark=star,only marks, xshift=-3.2mm, point meta=explicit symbolic] plot[error bars/.cd, y dir=both,y explicit] coordinates {(door,77.8)[ ] (fridge,81.2)[ ]};
\addplot+[black,mark=star,only marks, xshift=3.2mm, point meta=explicit symbolic] plot[error bars/.cd, y dir=both,y explicit] coordinates {(door,43.0)[ ] (fridge,62.6)[ ]};
\legend{Hand Guidance\qquad,\ourName{},Expert}
\end{axis}
\end{tikzpicture}
\vspace{-0.35cm}
\caption{Average completion times of new users. \myworries{Bars indicate standard errors.}\looseness=-1}
\label{fig:user_study}
\vspace{-0.5cm}
\end{figure}
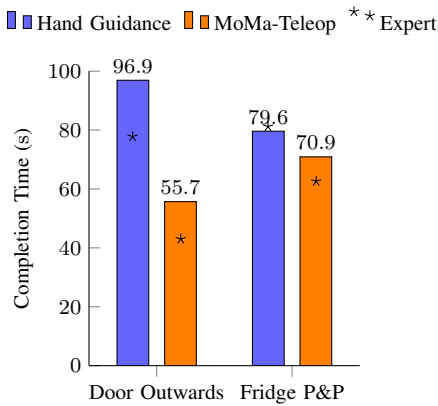

To evaluate the ease of use of the approaches for new users, we conduct a user study on the FMM robot for the \texttt{Door Outwards} and \texttt{Fridge P\&P} tasks.
We recruit six participants.
Each participant receives a short, five-minute introduction to each approach and is then given a practice attempt at the task. The user then completes three episodes for the best baseline, hand guidance, and our approach for each task. We change the order of the task and approach that each user starts with evenly and reverse the order of approaches for the second task. We instruct the users not to move the base while grasping an articulated object, as we found this to easily trigger the safety limits of the arm.
The results are reported in \cref{fig:user_study}.

We found large differences in user behaviors, strategies, and confidence. A particular challenge posed the understanding of joint limits, resulting in occasional failures with the arm joints locking for safety in both approaches, with an overall success rate of 91.7\% for both approaches.
The completion times confirm the relative results of an expert user. We find particularly large differences in the door opening task, which requires to follow specific motions over the large opening radius and, as a result, repeated base repositioning without a mobile base. Differences in the fridge task are less pronounced. As the fridge door can be opened from a static position, efficient base placement can complete the task with a single repositioning. However, even in such a more static task, users achieved an efficiency improvement of over 12\% with our approach\myworries{ (though not reaching statistical significance)}. One user found a particularly efficient strategy, outperforming the expert in both approaches on the fridge task, taking \SI{34.3}{\s} with hand guidance and \SI{25.7}{\s} with \ourName{}, even pulling the user average below the expert value. Overall, \ourName{} reduced average completion time by almost 40\%.\looseness=-1

\section{Imitation Learning}\label{sec:imitation}

\begin{table*}
    \centering
    \setlength{\tabcolsep}{4pt}
    \caption{Success rates of imitation learning policies from teleoperation data.}
    \vspace{-0.5em}
    \begin{threeparttable}
    \begin{tabularx}{.80\linewidth}{lcccccc}
    \toprule
     & &  Door Outwards & \multicolumn{2}{c}{Drawer} &  \multicolumn{2}{c}{Clean Table} \\
     \cmidrule(lr){3-3}
     \cmidrule(lr){4-5}
     \cmidrule(lr){6-7}
     Data Collection \myworries{Policy} & \myworries{Imitation} Policy & Unchanged & Unchanged & New Height & Unchanged & Obstacle \\
    \midrule
      Hand Guidance + Joystick & Whole-Body \myworries{Imitation}& \phantom{0}0 & 90 & n.e. & 0/90$^*$ & n.e.\\
      Hand Guidance + Joystick & EE \myworries{+ \ntwo{}} & \phantom{0}0 & 90 & n.e. & 0\myworries{/}90$^*$ & n.e. \\
     \midrule
      \ourName{} & Whole-Body \myworries{Imitation}& 80 & 100 & \phantom{0}0 & 90 & \phantom{0}0\\
      \ourName{} & EE \myworries{+ \ntwo{}} & 90 & 100 & 80 & 90 & 90\\
    \bottomrule
    \end{tabularx}
     \begin{tablenotes}[para,flushleft]
       \footnotesize      
       Unchanged: identical setup as for data collection. Obstacle: new obstacles added to the setting. New height: object placed at different height. n.e.: not evaluated. $^*$: depending data consistency, cf. \cref{sec:data_quality}.
     \end{tablenotes}
   \end{threeparttable}
    \label{tab:imitation_learning}
    \vspace{-0.5cm}
\end{table*}

\myworries{We then learn autonomous motions from} the collected demonstrations, \myworries{by leveraging} TAPAS-GMM~\cite{vonhartz2024imitation}, a state-of-the-art imitation learning method based on Gaussian Mixture Models (GMM). We use the time-based variant of TAPAS-GMM, which models the gripper action and end-effector pose \(ee_{t}\) across time in multiple task-relevant coordinate frames. To this end, TAPAS-GMM first segments long-horizon tasks, such as \textit{open the drawer}, into a series of shorter skills, such as \textit{grasp the handle} and \textit{pull the handle}. It then uses DINO features to extract a set of object keypoints~\cite{vonhartz2023treachery} from the robot's Intel RealSense D435 wrist camera. Subsequently, it automatically selects the relevant keypoints per skill and fits the set of demonstrations from the perspective of coordinate frames attached to the selected keypoints.
During inference, these per-frame models are joined using the current keypoint poses to generate a combined model in the world frame.
We then predict a full trajectory and step through it as long as the current end-effector pose is close enough to the last prediction. Otherwise, we repeat the last pose command.

We construct two \myworries{imitation} policies:
\textit{Whole-Body \myworries{Imitation}} jointly fits the GMM to the recorded end-effector and base poses and uses inverse kinematics to solve for arm and torso joint position commands while tracking the base and end-effector motions.
\textit{EE \myworries{+ \ntwo{}}} only models the gripper action, and end-effector poses \(ee_{t}\) and uses the same learned N$^2$M$^2$ base agent to convert the learned end-effector motions to whole-body motions. \myworries{We provide a detailed overview of the policies in the supplementary material.}

This system enables us to rapidly learn new mobile manipulation tasks from only \textit{five demonstrations}.
The combined data collection with \ourName{} and fitting of the models with TAPAS-GMM takes \textit{less than ten minutes} in total.

\subsection{Data Quality}\label{sec:data_quality}
\myworries{To evaluate the quality of the data generated by our approach, we collect five demonstrations with both the \textit{Hand Guidance + Joystick} and \textit{\ourName{}} methods across three tasks with hand guidance on the FMM robot: \texttt{Clean Table}, \texttt{Door Outwards} and an easily movable \texttt{Open Drawer} task. Then, we execute both imitation policies for ten episodes per task.} The results are presented in \cref{tab:imitation_learning}.\looseness=-1

We find that we can learn robust motions from static hand guidance data for tasks where a consistent teleoperation strategy in terms of order of base and arm execution exists, such as \texttt{Open Drawer} or \texttt{Clean Table}. For tasks that require large base motions and repositioning, the resulting trajectories are more complex and exhibit greater variance. For \texttt{Door Outwards}, the handle needs to be released and the base repositioned, which happens at different times and positions for different trajectories, rendering the trajectories difficult to model.
Consequently, end-effector, gripper, and base actions are not temporally aligned across trajectories, making the policy mix up parts of the motions due to the more complex data distribution.
Accurately fitting such data would require significantly more demonstrations.
We experienced the same issue in \texttt{Clean Table} when collecting data as a standard user would without first deciding on a consistent base positioning strategy.
This \myworries{more complex data distribution} resulted in a policy that is not sufficiently following the desired trajectory and struggling to coordinate base and end-effector\myworries{, failing to complete the task. We then repeated the experiment, with the operator first deciding on a consistent base placement. With this data, the imitation policy succeeded, but this requires both planning ahead and system knowledge}.

In contrast, the data from \ourName{} leads to smooth and consistent end-effector motions independent of the teleoperator's proficiency, as it removes the decision about base placements and allows the completion of mobile manipulation motions without regrasping.
Using its data, we are able to learn both successful pure end-effector motions as well as whole-body motions from very few demonstrations due to the reduced coordination effort required from the end-effector policy. Its data resulted in shorter trajectories and lower execution times for both policies, as the end-effector motions are always focusing on the task, while the static hand guidance data results in unnecessary end-effector movements from when the arm is idle on top of the moving base.
The remaining failures stem mostly from the accumulated noise of depth sensors, keypoint estimation, and whole-body motions, resulting in insufficiently precise grasping.

\subsection{Generalization \myworries{to New Contexts}}\label{sec:generalization}
We further evaluate the policies' ability to generalize to new contexts. The keypoints and task-parameterized motions are object-centric, enabling direct transfer to different positioning of the objects. As such, the learned end-effector motions transfer directly to new contexts, with the base agent enabling the kinematic feasibility of the trajectory.
In contrast, the whole-body policy jointly models the base motions and end-effector motions.
Consequently, they are mutually dependent, for example, due to the kinematic limits of the robot.
Thus, they do not easily generalize to new contexts, such as a changed height of the drawer.
Similarly, new obstacles would require the whole-body model to learn simultaneous obstacle avoidance across a wide range of different obstacle configurations. As such, both components would require a lot of additional training data.

To evaluate this, we adapt the tasks with common scenarios, as they might occur in a household: we place the drawer at a different height and add a new obstacle at three different positions in front of the table to clean. The scenarios are shown in the supplementary, and the results are shown in \cref{tab:imitation_learning}. 
We find that the whole-body policy does not generalize to these scenarios, failing to reach the required end-effector poses from the learned base movement and colliding with the obstacles. In contrast, the EE policy directly adapts to these scenarios, with no drop in performance.
\section{Conclusion}
\myworries{We introduced a novel teleoperation approach that translates 6 DoF user signals to end-effector motions, which are then subsequently converted to compliant whole-body motions by a reinforcement learning agent. This enables whole-body teleoperation from existing, highly affordable control modalities, in contrast to existing methods that require expensive and specialized exoskeletons and puppeteers or tracking approaches that restrict the operator.}
\myworries{We then demonstrated how this system can be integrated with recent task-parameterized GMMs}
for autonomous execution of the learned tasks, \myworries{and} generalization to new situations from as little as five demonstrations. 
We made the code publicly available to facilitate \myworries{the collection of large-scale mobile manipulation datasets and rapid skill learning across a wide range of tasks, including contact-rich manipulation.}

\bibliographystyle{IEEEtran}
\bibliography{root.bib}

\clearpage
\renewcommand{\baselinestretch}{1}
\setlength{\belowcaptionskip}{0pt}

\begin{strip}
\begin{center}
\vspace{-5ex}
\textbf{\LARGE \bf
\ourtitle{}
} \\
\vspace{3ex}

\Large{\bf- Supplementary Material -}\\
\vspace{0.4cm}
\normalsize{Daniel Honerkamp$^{*}$, Harsh Mahesheka$^{*}$, Jan Ole von Hartz, Tim Welschehold and Abhinav Valada}\\
\end{center}
\end{strip}

\setcounter{section}{0}
\setcounter{equation}{0}
\setcounter{figure}{0}
\setcounter{table}{0}
\setcounter{page}{1}
\makeatletter

\renewcommand{\thesection}{S.\arabic{section}}
\renewcommand{\thesubsection}{S.\arabic{subsection}}
\renewcommand{\thetable}{S.\arabic{table}}
\renewcommand{\thefigure}{S.\arabic{figure}}

\let\thefootnote\relax\footnote{$^*$These authors contributed equally. All authors are with the Department of Computer Science, University of Freiburg, Germany.\\
Project page: \website{}
}%
\normalsize 
In this supplementary material, we provide details on the comparison criteria to existing approaches, the joystick teleoperation configurations, the evaluated tasks, the tracking workspaces, and failure cases. Moreover, we provide an overview of the different policies used across this work and report statistical significance of the teleoperation experiments. Demonstrations of teleoperation with all approaches are provided in the accompanying video and the project page.

\subsection{Comparison Criteria}\label{app:comparison_criteria}
We define the following criteria for the comparison of existing mobile manipulation teleoperation approaches. The categories Cost, Modality, Height Control, Whole-Body Teleoperation, Robot Agnostic, and Action Space are based on the definitions in~\citeS{dass2024telemoma}, with some adaptations or extensions. In particular, we add an additional cost category.
\begin{itemize}[topsep=0pt]
    \item \textit{Cost:}\\ \\
        \begin{tabular}{cl}
         \textbf{\$}: & \$0 -- 100 (Joysticks, Kinesthetic w/o\\
                      & extra sensors)\\
         \textbf{\$\$}: & \$100 -- 1,000 (VR, Vision, Phone, \\
                        & Kinesthetic with additional F/E sensors)\\ 
         \textbf{\$\$\$}: & \$1,000 -- 10,000 (Mocap Systems) \\ 
         \textbf{\$\$\$\$}: & \$10,000+ (Custom Hardware)\\ \\
        \end{tabular}
    \item \textit{Modality:} the human interface used by the human operator for teleoperation (e.g. virtual reality (VR), puppeteering with a kinematically similar device, motion capture systems (Mocap), etc.).
    \item \textit{Workspace:} The space within which the human operator can move and control the robot. This may impose restrictions on how far it is possible to move and whether the operator can observe the robot from close by when executing high-precision actions such as grasping a handle. ``Tracked space" denotes the requirement to stay within tracked space or field of view of a Mocap system, VR system, or a tracking RGBD camera. ``Unlimited" denotes no restrictions.
    \item \emph{Height Control:} True if the paper demonstrates control of the robot's torso joint. 
    \item \emph{Whole-Body Teleoperation:} True if simultaneous arm and base motion is enabled by the method. 
    \item \emph{Robot Agnostic:} True if the method works for many different robots; false if it is specific to a particular platform. 
    \item \emph{Action Space:} ``EE Pose(s)" denotes control of the robot's end-effector(s) in Cartesian space, whereas ``Joint Pos." indicates joint-space control for the arms and/or torso. Base Vel. indicates control of the base velocity; TRILL~\citeS{seo2023deep} allows users to select among predefined gaits with a VR controller, denoted ``Gait". MOMA-Force~\citeS{yang2023moma} enables teleoperation of end-effector Cartesian pose through kinesthetic teaching, and additionally, records desired end-effector wrenches, denoted ``EE Pose and Wrench". TeleMoMa~\citeS{dass2024telemoma} allows users to control end-effector Cartesian pose, base velocity, and torso joint position; In Zhao~et~al.~\citeS{zhao2022hybrid}, the user guides the end-effector and switches the loco-manipulation mode between base and end-effector. \ourName{} reduces the action space for the operator to pure end-effector poses but converts these to whole-body motions via the base agent.
    \item \textit{Wrench Data}: True if the approach is capable of demonstrating precise wrench values by the end-effector or robot joints, such as through physical guidance or with a portable end-effector with corresponding sensors.
    \item \textit{Obstacle Avoidance}: ``Manual (M)" means the human operator is responsible for issuing commands that avoid any obstacles. ``Autonomous (A)" means that the system autonomously avoids obstacles. False if the work does not integrate or demonstrate any obstacle avoidance.
\end{itemize}

\subsection{Joytick Configurations}\label{app:joystick}
\setlength{\tabcolsep}{1pt}
\begin{figure*}[t]
	\centering
	\resizebox{\linewidth}{!}{%
 \includegraphics[width=\linewidth,trim={0cm 0cm 0cm 0cm},clip,angle =0,valign=c]{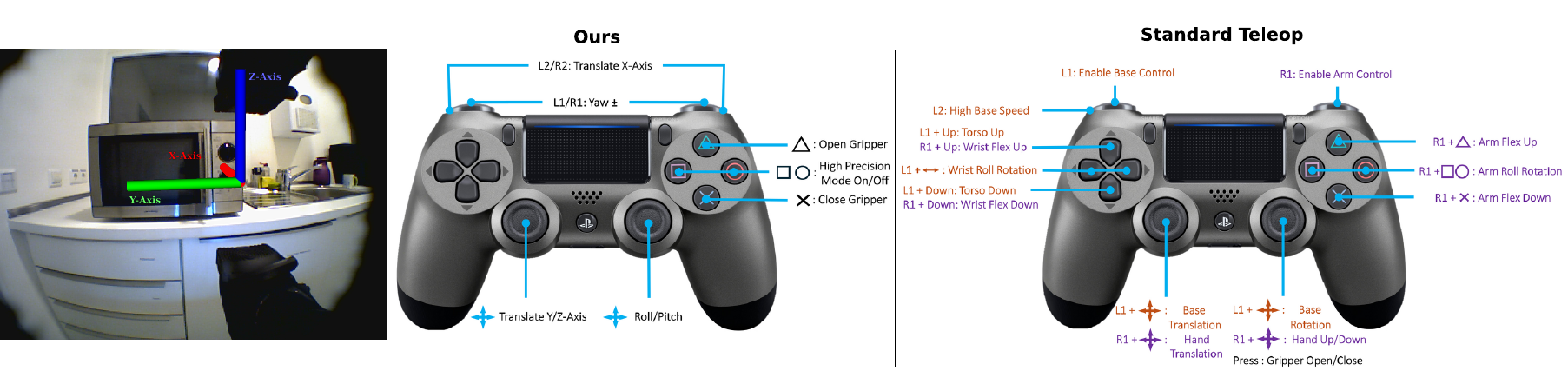}}
	 \vspace{-0.2cm}
     \caption{Left: Reference frame for the control inputs in the wrist camera view of the HSR robot and button assignment of \ourName{}. Right: Button assignment of the original teleoperation ROS package developed for the HSR.} 
  	\label{fig:joystick}
\end{figure*}
\setlength{\tabcolsep}{6pt}
\cref{fig:joystick} shows the joystick configuration for our approach as well as the baseline. The commands for \ourName{} are issued in the frame of the wrist camera, shown on the left. The sticks and shoulder buttons then control the translation and orientation of the end-effector in this frame. Two additional buttons enable grasping and activation of the high-precision mode, as shown in the middle. In contrast, the original teleoperation approach developed for the robot requires the user to use the shoulder buttons to toggle between control modes for the arm and base, having to overload buttons to achieve full control depending on whether the L1 or R1 button is held down, the meaning of the buttons changes. In our experiments, we found that this risks to confuse different buttons.

\subsection{Task descriptions}\label{app:tasks}
\begin{figure*}[htb]
    \centering
    \captionsetup[subfigure]{justification=centering}
    \subfloat[\texttt{Open Drawer}\\ \vspace{0.5em}\footnotesize{Placement during data collection.\looseness=-1}]{\includegraphics[height=1.28in]{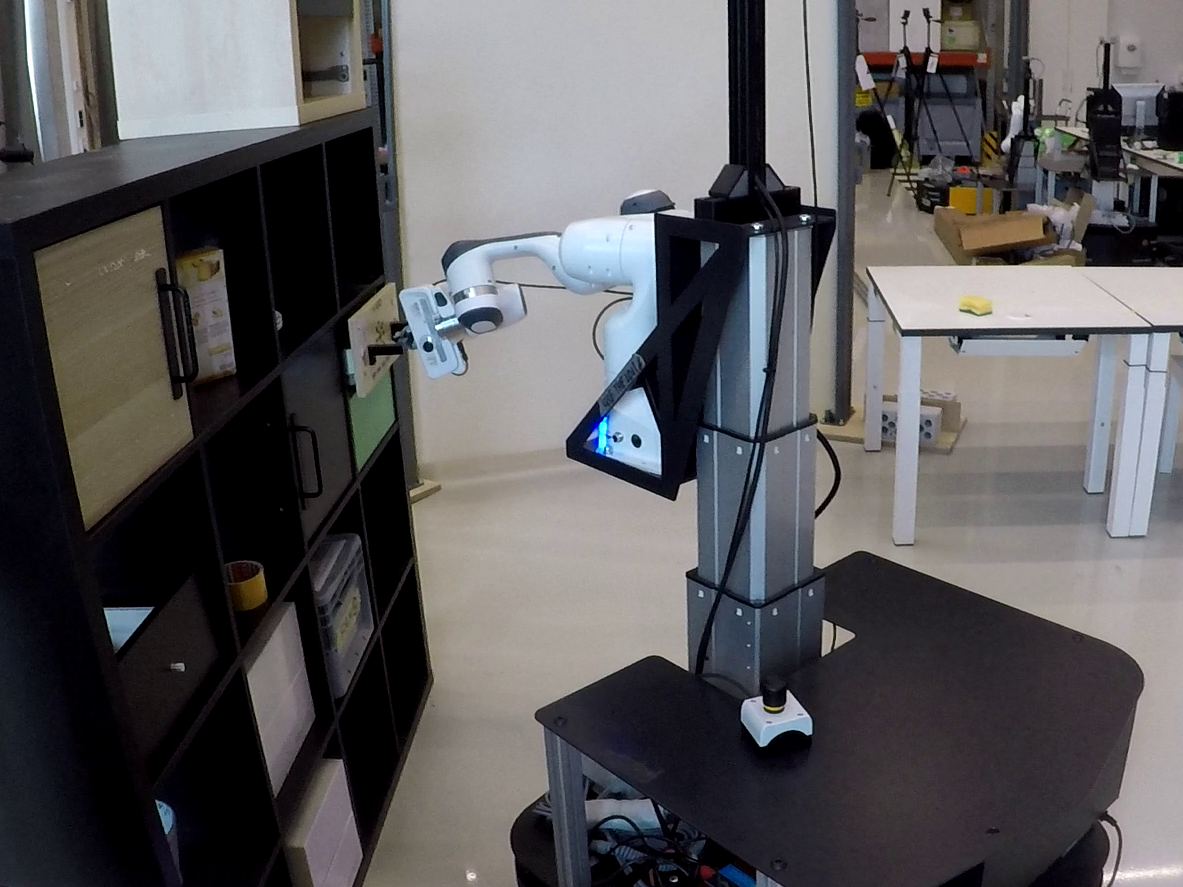}}\hspace{0.25pt}
    \subfloat[\texttt{Open Drawer High}\\ \vspace{0.5em}\footnotesize{Changed scenario at new height.}]{\includegraphics[height=1.28in]{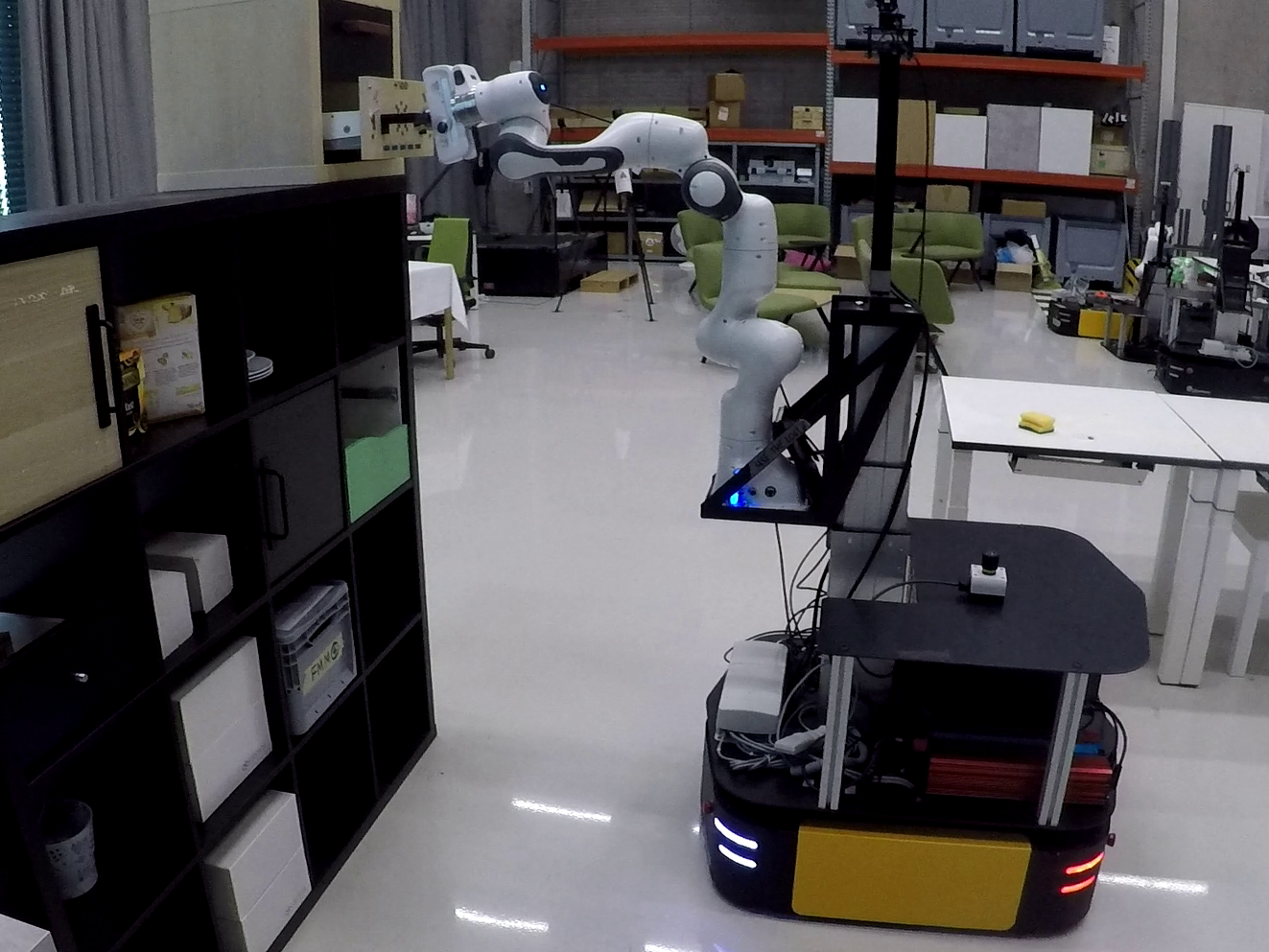}}\\
    \subfloat[\texttt{Clean Table}\\ \vspace{0.5em}\footnotesize{First new obstacle position.}]{\includegraphics[height=1.28in]{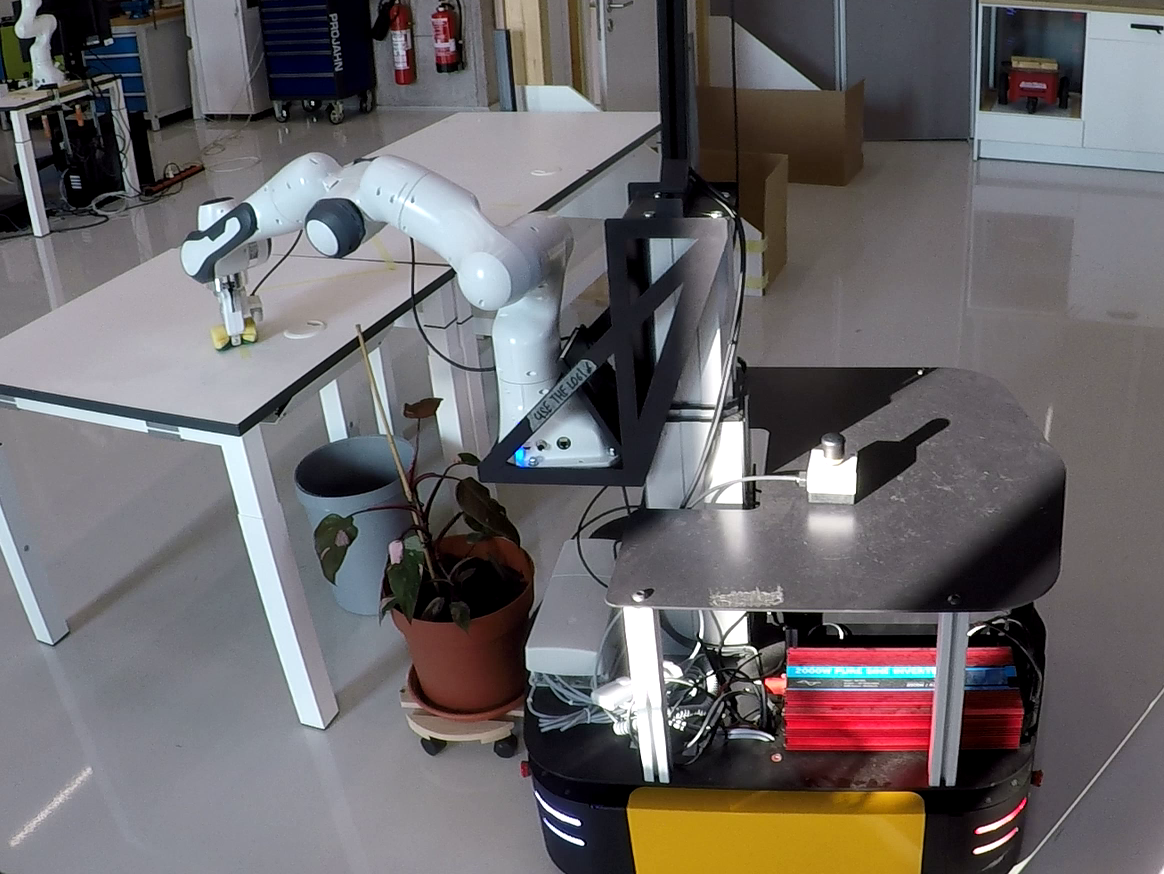}}\hspace{0.25pt}
    \subfloat[\texttt{Clean Table}\\ \vspace{0.5em}\footnotesize{Second new obstacle position.}]{\includegraphics[height=1.28in]{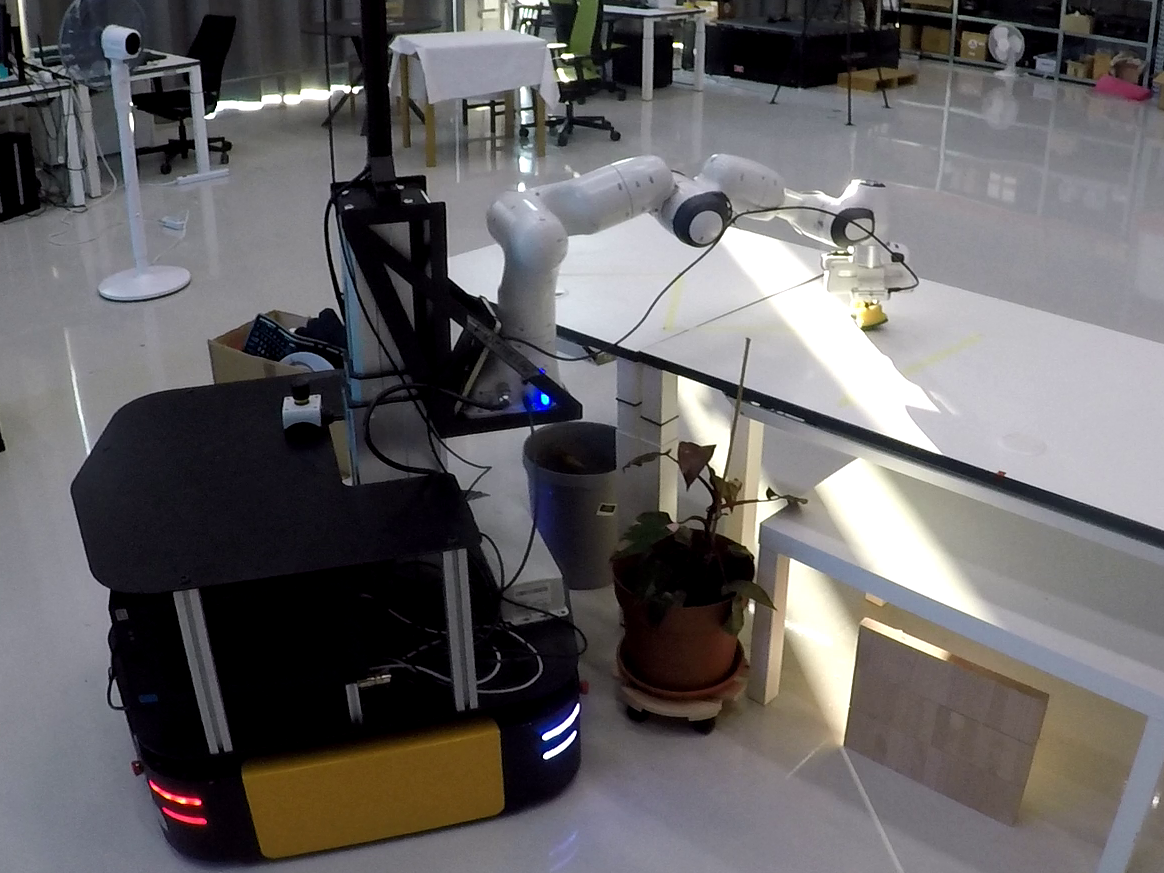}}\hspace{0.25pt}
    \subfloat[\texttt{Clean Table}\\ \vspace{0.5em}\footnotesize{Third new obstacle position.}]{\includegraphics[height=1.28in]{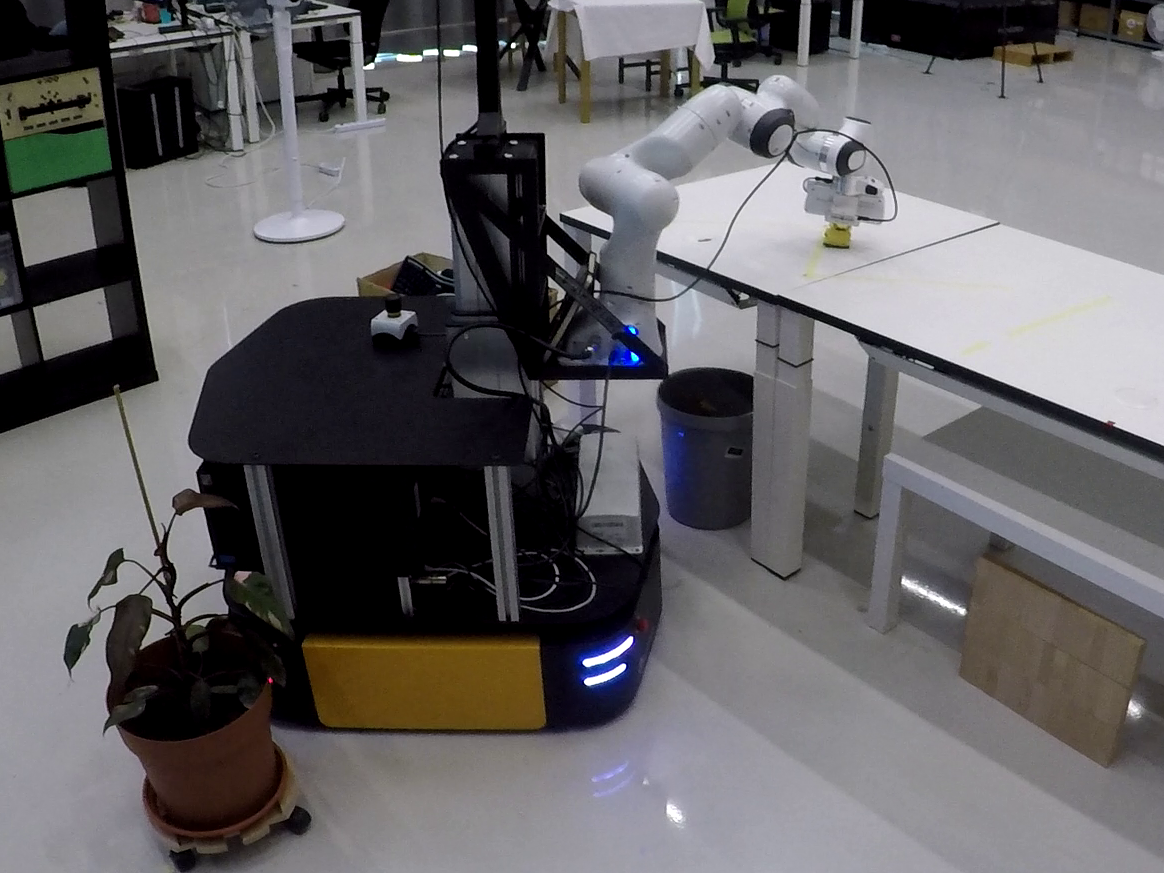}}
     \caption{Task scenarios evaluated for imitation learning.}\label{fig:tasks_imitation}
\end{figure*}

\noindent{\texttt{Microwave:}} The robot has to open a microwave in a narrow office kitchen.

\noindent{\texttt{Door Inwards:}} Open a door inwards using the door handle while driving through the frame. As the HSR does not have enough strength for the latch, we disable the spring in the handle.

\noindent{\texttt{Toolbox:}} open a toolbox with a rotational joint upwards. The box is initially unlatched as the latches cannot be operated with a parallel gripper.

\noindent{\texttt{P\&P:}} grasp a bottle from a small coffee table and place it on top of a high shelf. 

\noindent{\texttt{Folding Cabinet:}} open a cabinet with an upwards-folding door. 

\noindent{\texttt{Door Outwards:}} unlatch the handle and open the door outwards. During imitation learning, we disable the hatching mechanism as its strong spring frequently triggers safety violations of the Franka Arm.

\noindent{\texttt{Clean Table:}} equipped with a sponge in the end-effector, clean a table by scrubbing along a given path (marked by tape). During imitation learning, the sponge is placed at the beginning of the line to provide keypoint references.

\noindent{\texttt{Fridge P\&P}}: open a fridge, grasp a carton of milk out of the door of the fridge, and place it down on a small shelf next to it.

For imitation learning, we introduce an additional \texttt{Open Drawer} task and introduce unseen scenarios. These tasks are shown in \cref{fig:tasks_imitation}. For the obstacles, we evaluate over 3 / 3 / 4 episodes per position, matching the total of ten episodes for each task.

\setlength{\tabcolsep}{1pt}
\renewcommand{\arraystretch}{1}
\begin{figure*}
	\centering
    \footnotesize
    \setlength{\tabcolsep}{0.0cm}%
    {\renewcommand{\arraystretch}{1}%
 \includegraphics[width=.8\linewidth,trim={0cm 0cm 0cm 0cm},clip,angle =0,valign=c]{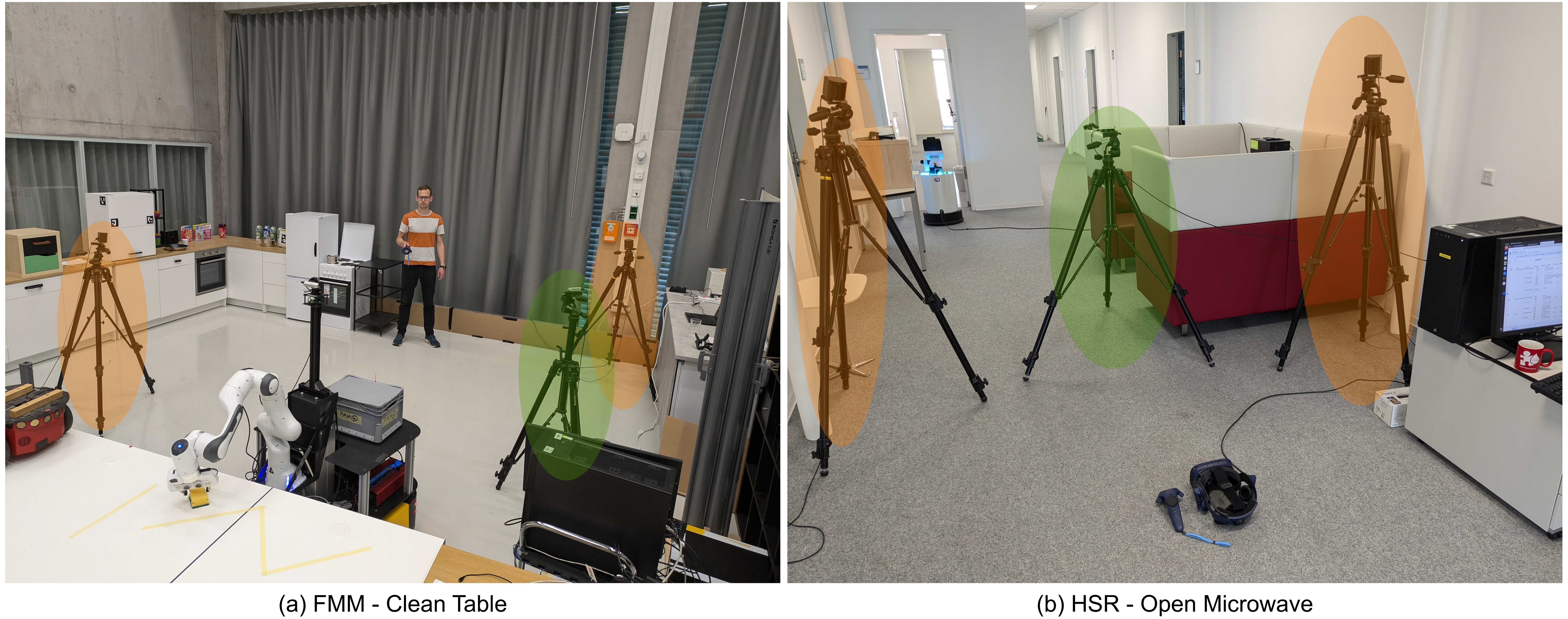}}
	\caption{Workspace setup for the tracking methods. The Vision tracking method requires one camera stand with an RGB-D camera (marked green). The VR tracking method additionally requires (at least) two lighthouses (marked orange). (a) FMM robot performing the clean table task. (b) HSR robot for performing the microwave task in the narrow office kitchen.}
  	\label{fig:workspaces}
\end{figure*}
\setlength{\tabcolsep}{6pt}
\renewcommand{\arraystretch}{1}

\setlength{\tabcolsep}{1pt}
\renewcommand{\arraystretch}{1}
\begin{figure*}
	\centering
    \footnotesize
    \setlength{\tabcolsep}{0.0cm}%
    {\renewcommand{\arraystretch}{1}%
 \includegraphics[width=\linewidth,trim={0cm 0cm 0cm 0cm},clip,angle =0,valign=c]{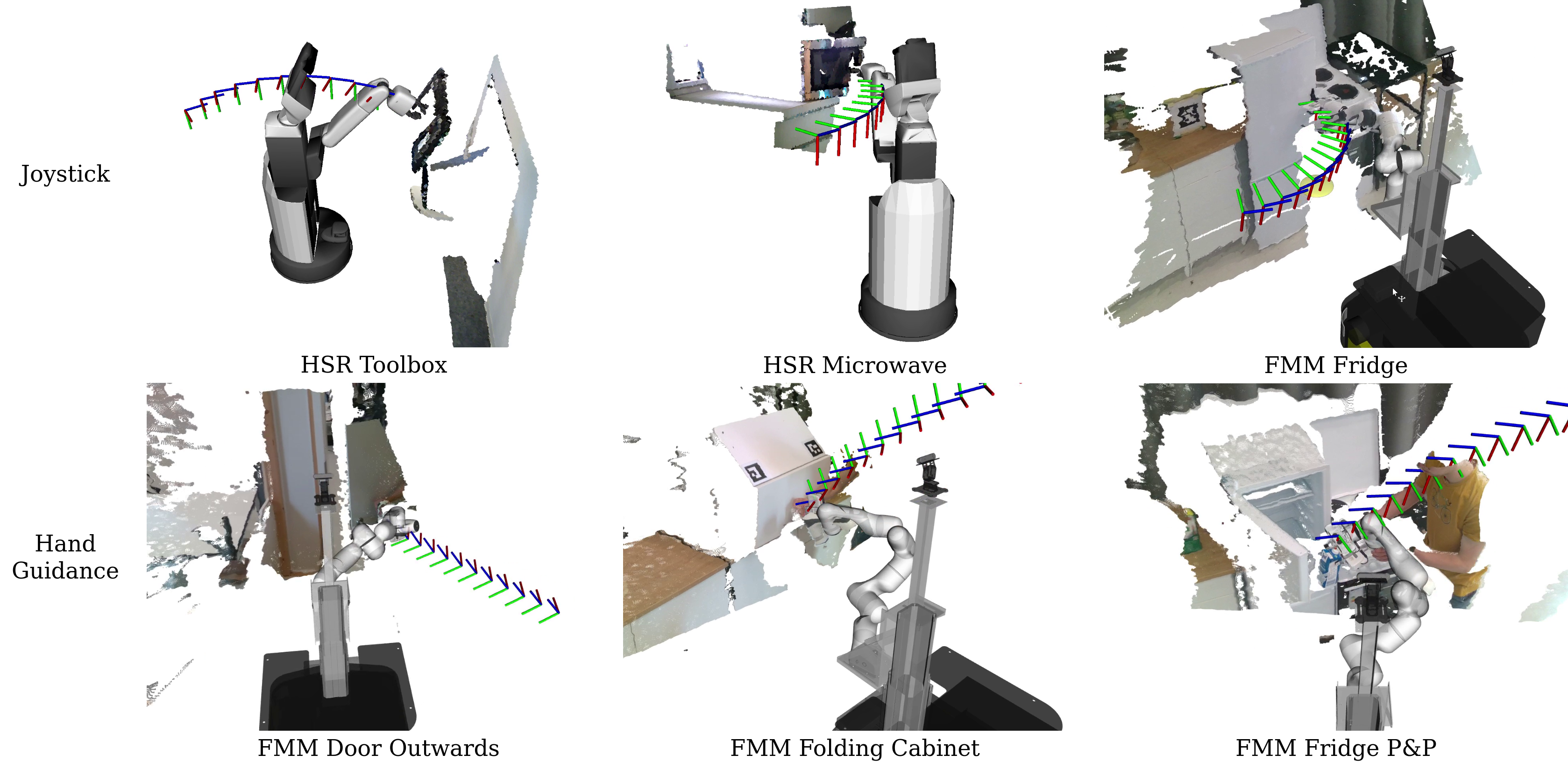}}
	\caption{End-effector motions inferred from joystick signals (top) and hand guidance (bottom) across different tasks.}
  	\label{fig:eemotions}
\end{figure*}
\setlength{\tabcolsep}{6pt}
\renewcommand{\arraystretch}{1}
\subsection{End-effector Motions}\label{app:ee-motions}
\cref{fig:eemotions} shows the end-effector motions inferred from the user signals across different tasks and input interfaces.
We experimentally evaluated alternative functional forms, in particular, the direct fitting of non-linear regression through the history of end-effector poses in hand guidance mode. However, we found this process unreliable, as the length of the history and the assumptions on the functional form of the curve required a lot of tuning and showed to be very task-dependent.

\subsection{Tracking workspaces}
\cref{fig:workspaces} shows the setup for the tracking baselines. The requirement for up to three camera stands together with ample room for the operator to move as much as the robot has to move results in a significant distance between the operator and the robot. The robot itself blocking the view of the end-effector or task-relevant objects means additional difficulties in observing the task closely.

\setlength{\tabcolsep}{1pt}
\begin{figure*}[t]
	\centering
	\resizebox{.9\linewidth}{!}{%
 \includegraphics[width=\linewidth,trim={0cm 0cm 0cm 0cm},clip,angle =0,valign=c]{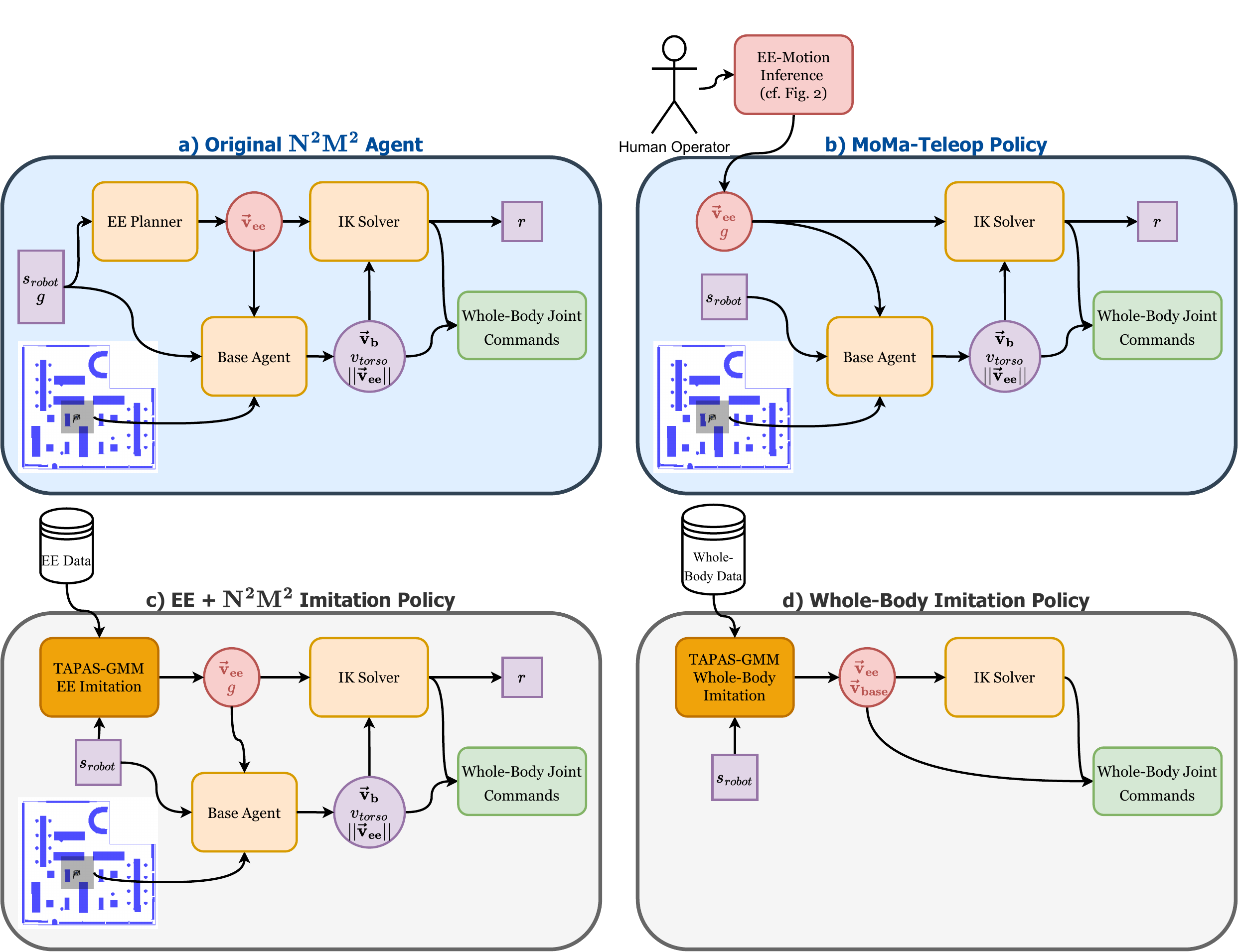}}
     \caption{\myworries{Comparison of the different policies. a) Original \ntwo{} agent as developed in~\cite{honerkamp2023learning}. b) Integration of MoMa-Teleop with the \ntwo{} agent with motions inferred from a human operator (cf. Fig. 2 for the complete system). c) Imitation of the collected end-effector motions, with reuse of the base agent. d) Full imitation of the collected whole-body motions.}} 
  	\label{fig:policy_overview}
\vspace{-0.5cm}
\end{figure*}
\setlength{\tabcolsep}{6pt}

\subsection{Tracking failure cases}\label{sec:tracking_failures}
\para{Limited operator workspace:} The field of view of either the camera or VR lighthouses limits the spatial extent for mobile manipulation tasks and requires careful initial positioning of the operator to have enough space in the directions required for the task. The workspace setups are shown in \cref{fig:workspaces}.\\
\para{Distant operation:} Existing environments do not always provide enough space to set up the workspace next to the task. As a result, the robot can only be watched from a distance or be operated remotely through a camera (adding latency). If positioned behind the robot, the robot itself may occlude handles or other task-relevant parts from the operator. This makes precise motions such as grasping harder.\\
\para{Embodiment mismatch:} If the embodiments differ strongly, torso movements of the operator can result in largely different inverse kinematics solutions and, as a result, fast, unwanted arm motions. For arms with a higher degree of freedom, it is furthermore challenging to understand good base and relative end-effector placements for certain tasks. For example, should the FMM robot position its base right in front of or orthogonal to a cabinet to reach a handle at a low height? This can result in unstable inverse kinematics solutions and, as a result, imprecise or fast arm movements when reaching the edge of the workspace. \\
\para{Rotation:} For vision tracking, turning \SI{90}{\degree} or more resulted in failure to accurately detect the hand orientation as the palm of the hand moves out of view. VR Tracking can support larger orientation changes but at the cost of additional lighthouses.

For \texttt{Toolbox}, VR Tracking repeatedly pushed down the handle (requiring human intervention to put it back up - not considered a failure). When grasping, it was not possible to pull in the required direction without pulling the heavy toolbox around.
For the FMM robot, we find the VR Tracking approach to be able to track the pattern for \texttt{Clean Table} roughly, though with large deviations. Additionally, the operator was unable to keep a constant pressure on the table. In contrast, hand guidance enables the demonstrator to produce a desired level of pressure by directly guiding the hand physically.
On the \texttt{Door Outwards} task, we found the FMM unable to unlatch the door handle. At low stiffness settings, slipping off, while at high stiffness settings, triggering safety violations. We then attempt to open the door without latching the handle. In this case, the arm repeatedly either collided with the tower of the robot, the low stiffness masked the wrenches acting on the arm until it slips off and rebounds or safety stops are triggered when reaching joint limits, as the simultaneous base and arm motions result in too much force on the arm.

Vision Tracking additionally struggled with a missing safety stop, requiring a second person to stop tracking. Torso control can require the operator to squad down for prolonged periods, which can be difficult to hold.

\begin{table*}
\color{myworriestablecolor}
    \centering
    \caption{Teleoperation results with standard errors across robots and tasks.} 
    \vspace{-0.5em}
    \setlength{\tabcolsep}{2pt}
    \begin{threeparttable}
    \begin{tabularx}{\textwidth}{lcccccccccc}
    \toprule
      \multicolumn{1}{c}{\textbf{HSR Robot}} & \multicolumn{2}{c}{P\&P} & \multicolumn{2}{c}{Microwave} & \multicolumn{2}{c}{Door Inwards} & \multicolumn{2}{c}{Toolbox} & \multicolumn{2}{c}{Average} \\
      \cmidrule(lr){2-3}
      \cmidrule(lr){4-5}
      \cmidrule(lr){6-7}
      \cmidrule(lr){8-9}
      \cmidrule(lr){10-11}
            Model & SR & Time & SR & Time & SR & Time & SR & Time & SR & Time \\
      \midrule
      Joystick & $100 \pm 0.00$ & $42.0 \pm 0.55$& $100 \pm 0.00$& $42.0 \pm 1.41$ & $100 \pm 0.00$ & $66.8 \pm 4.43$ & $100 \pm 0.0$& $83.6 \pm 4.57$ & $100 \pm 0.00$& 58.6\\
      Vision Tracking & $\phantom{0}40 \pm 21.9$ & $41.0 \pm 3.00$ & $\phantom{0}60 \pm 21.9$ & $43.7 \pm 9.49$ & n.e. & n.e. & n.e. & n.e.  & $\phantom{0}25 \pm 9.7$ & (42.4)\\
      VR Tracking & $100 \pm 0.00$& $38.4 \pm 3.31$& $\phantom{0}80 \pm 17.9$ & $46.5 \pm 7.24$ & 0/($80^* \pm 17.9$) & ($84.5^* \pm 18.66$) & $\phantom{00}0 \pm 0.0$ & --  & $\phantom{0}45 \pm 11.1$ & (42.5) \\
      \ourName{} & $100 \pm 0.00$& $44.2 \pm 1.98$ & $100 \pm 0.00$ & $36.2 \pm 1.80$ & $80 \pm 17.9$ & $46.3 \pm 2.25$ & $100 \pm 0.0$& $55.2 \pm 4.61$ & $\phantom{0}95 \pm 0.05$& 45.5\\
    \midrule
    \midrule
      \multicolumn{1}{c}{\textbf{FMM Robot}} & \multicolumn{2}{c}{Clean Table} & \multicolumn{2}{c}{Door Outwards} & \multicolumn{2}{c}{Folding Cabinet} & \multicolumn{2}{c}{Fridge P\&P} & \multicolumn{2}{c}{Average} \\
      \cmidrule(lr){2-3}
      \cmidrule(lr){4-5}
      \cmidrule(lr){6-7}
      \cmidrule(lr){8-9}
      \cmidrule(lr){10-11}
            Model & SR & Time & SR & Time & SR & Time & SR & Time & SR & Time \\
            \midrule
    Hand Guidance & $100 \pm 0.00$& $42.8 \pm 1.43$ & $\phantom{0}80 \pm 17.9$ & $77.8 \pm 2.17$ & $80 \pm 17.9$ & $62.5 \pm 17.56$ & $100 \pm 0.0$& $81.2 \pm 4.89$ & $\phantom{0}90 \pm 6.7$ & 66.1 \\
    Vision Tracking & n.e. & n.e. & n.e. & n.e. & n.e. & n.e. & n.e. & n.e. & n.e. & n.e.\\
    VR Tracking & $100 \pm 0.00$& $114 \pm 15.64$ & $\phantom{00}0 \pm 0.00$ & -- & n.e. & n.e. & n.e. & n.e. & $\phantom{0}25 \pm 9.7$ & (114) \\
    \ourName{} & $100 \pm 0.00$& $38.4 \pm 0.87$& $100 \pm 0.00$ & $43.0 \pm 3.70$ & $80 \pm 17.9$ & $43.3 \pm 5.25$& $100 \pm 0.0$& $62.6 \pm 4.19$& $\phantom{0}95 \pm 4.9$ & 46.8\\
    \bottomrule
    \end{tabularx}
        \begin{tablenotes}[para,flushleft]
       \footnotesize      
       SR: average success rate in percent, time: average completion time in seconds over the \textit{successful} attempts, n.e.: not evaluated on this task due to hardware safety concerns, $^*$: finished opening door, but was unable to grasp and follow the handle of the door. $\pm$ reports estimated standard errors.
     \end{tablenotes}
   \end{threeparttable}
    \label{tab:teleop_significance}
\end{table*}
\begin{table*}
\color{myworriestablecolor}
    \centering
    \caption{Statistical signifance levels. }
    \label{tab:significance}
    \begin{threeparttable}
    \begin{tabularx}{.85\linewidth}{llr}
        \toprule
        \textbf{Metric} & \textbf{Null Hypothesis} & \textbf{P-Value} \\
        \midrule
        Average SR on HSR Robot & No difference in proportions of \ourName{} vs Joystick & 1.0 \\
        Average SR on HSR Robot & No difference in proportions of \ourName{} vs Vision Tracking & 1.002e-05$^{***}$ \\
        Average SR on HSR Robot & No difference in proportions of \ourName{} vs VR Tracking & 0.0012$^{***}$\\
        Average SR on FMM Robot & No difference in proportions of \ourName{} vs Hand Guidance & 1.0 \\
        Average SR on FMM Robot & No difference in proportions of \ourName{} vs VR Tracking & 1.002e-05$^{***}$\\      
        \midrule
        Average Duration P\&P & The difference of the means is zero for \ourName{} vs Joystick & 0.317 \\
        Average Duration Microwave & The difference of the means is zero for \ourName{} vs Joystick &0.034$^{**}$\\
        Average Duration Door Inwards & The difference of the means is zero for \ourName{} vs Joystick & 0.007$^{***}$\\
        Average Duration Toolbox & The difference of the means is zero for \ourName{} vs Joystick & 0.002$^{***}$\\
        \midrule
        Average Duration Clean Table & The difference of the means is zero for \ourName{} vs Hand Guidance & 0.030$^{**}$\\
        Average Duration Door Outwards & The difference of the means is zero for \ourName{} vs Hand Guidance & 1.332e-4$^{***}$\\
        Average Duration Folding Cabinet & The difference of the means is zero for \ourName{} vs Hand Guidance & 0.334\\
        Average Duration Fridge P\&P & The difference of the means is zero for \ourName{} vs Hand Guidance & 0.020$^{**}$\\
        \bottomrule
    \end{tabularx}
     \begin{tablenotes}[para,flushleft]
       \footnotesize      
       $^*$ significant at 90\% $^{**}$ significant at 95\% $^{***}$ significant at 99\%. Evaluated via Fisher's exact test for binary success rates (SR) and via two-sided Student's t-test for continuous durations.
     \end{tablenotes}
     \end{threeparttable}
\end{table*}

\subsection{\myworries{Policy Overview}}\label{sec:policies}
\myworries{\cref{fig:policy_overview} provides an overview of the different policies used in this work:}
\begin{itemize}
    \item[a)] \myworries{\textit{\ntwo{}:} The original \ntwo{} policy as developed in~\citeS{honerkamp2023learning}, relying on a given end-effector planner module, that can be swapped out at test time. This agent is trained with reinforcement learning to ensure the kinematic feasibility of given end-effector motions.}
    \item[b)] \myworries{\textit{\ourName{}:} our proposed approach to generate whole-body motions for operation of a robot by a human operator. Refer to \cref{sec:approach} for details on the inference and generation of the end-effector motions.}
    \item[c)] \myworries{\textit{EE + \ntwo{} Imitation Policy:} TAPAS-GMM based policy that models the end-effector motions from the collected data, then uses the same \ntwo{} agent to convert them to whole-body motions, as detailed in \cref{sec:imitation}.}
    \item[d)] \myworries{\textit{Whole-Body Imitation Policy:} TAPAS-GMM based policy that models both end-effector and base motions from the collected data and, together with inverse kinematics for the remaining joints, directly produces whole-body motions, as detailed in \cref{sec:imitation}.}
\end{itemize}

\subsection{\myworries{Statistical Significance}}
\myworries{
We report standard errors for all tasks according to \cref{eq:standard_errors} as a measure of the spread of the outcomes in \cref{tab:teleop_significance}. Note that we do not report standard errors for the average duration across tasks, as these are weighted averages of different numbers of successful episodes across the different approaches.

\begin{equation}\label{eq:standard_errors}
    SE = \frac{\hat{\sigma}}{\sqrt{n}}
\end{equation}

For the binary success outcomes, this becomes the Wald interval. Note that in this case, the intervals can suffer from overshoot, resulting in the 0.0 standard errors if all episodes were completed successfully~\citeS{newcombe1998two}.

We, therefore, first test the differences in success rates for statistical significance, using Fisher's exact test~\citeS{fisher1922interpretation}, which is more appropriate than the standard error intervals for binary outcomes with the given sample sizes. Due to the sample sizes, we cannot conclude that the differences in success rates on each individual task differ significantly. However, we can conclude that the Vision Tracking and VR Tracking approaches are able to complete significantly fewer tasks across each robot. The resulting p-values and levels of statistical significance are reported in \cref{tab:significance}.

In a second step, we compare the speed of the three approaches that were able to consistently solve all tasks: Joystick on the HSR, Hand Guidance on the FMM, and \ourName{}. For this, we evaluate the statistical significance of the average durations per task with a two-sided Student's t-test~\citeS{student1908probable}. The results are reported in \cref{tab:significance}. We can conclude that, for six out of the eight tasks, \ourName{} completes the task significantly faster than the baselines.
}

\clearpage

{\footnotesize
\bibliographystyleS{IEEEtran}
\bibliographyS{root.bib}
}

\end{document}